\documentclass[sigconf,screen]{acmart}

\usepackage[normalem]{ulem}
\usepackage{pervasives}

\AtBeginDocument{%
  \providecommand\BibTeX{{%
    \normalfont B\kern-0.5em{\scshape i\kern-0.25em b}\kern-0.8em\TeX}}}

\setcopyright{rightsretained}
\acmPrice{}
\acmDOI{10.1145/3582016.3582047}
\acmYear{2023}
\copyrightyear{2023}
\acmSubmissionID{asplosc23main-p163-p}
\acmISBN{978-1-4503-9918-0/23/03}
\acmConference[ASPLOS '23]{Proceedings of the 28th ACM International Conference on Architectural Support for Programming Languages and Operating Systems, Volume 3}{March 25--29, 2023}{Vancouver, BC, Canada}
\acmBooktitle{Proceedings of the 28th ACM International Conference on Architectural Support for Programming Languages and Operating Systems, Volume 3 (ASPLOS '23), March 25--29, 2023, Vancouver, BC, Canada}
\received{2022-07-07}
\received[revised]{2022-11-03}
\received[accepted]{2023-01-19}

\begin{document}

\author{Zihao Ye}
\authornote{Part of this work was done during internship at OctoML.}
\affiliation{
  \institution{University of Washington}
  \city{Seattle}
  \state{WA}
  \country{USA}
}
\email{zhye@cs.washington.edu}

\author{Ruihang Lai}
\authornote{Part of this work was done at Shanghai Jiao Tong University.}
\affiliation{
  \institution{Carnegie Mellon University}
  \city{Pittsburgh}
  \state{PA}
  \country{USA}
}
\email{ruihangl@cs.cmu.edu}

\author{Junru Shao}
\affiliation{
  \institution{OctoML}
  \city{Seattle}
  \state{WA}
  \country{USA}
}
\email{jshao@octoml.ai}

\author{Tianqi Chen}
\authornote{Also with OctoML.}
\affiliation{
  \institution{Carnegie Mellon University}
  \city{Pittsburgh}
  \state{PA}
  \country{USA}
}
\email{tqchen@cmu.edu}

\author{Luis Ceze}
\authornotemark[3]
\affiliation{
  \institution{University of Washington}
  \city{Seattle}
  \state{WA}
  \country{USA}
}
\email{luisceze@cs.washington.edu}

\title{\sys: \papertitle}

\date{}

\thispagestyle{empty}

\begin{abstract}

Sparse tensors are rapidly becoming critical components of modern deep learning workloads. However, developing high-performance sparse operators can be difficult and tedious, and existing vendor libraries cannot satisfy the escalating demands from new operators. Sparse tensor compilers simplify the development of operators, but efficient sparse compilation for deep learning remains challenging because a single sparse format cannot maximize hardware efficiency, and single-shot compilers cannot keep up with latest hardware and system advances.  In this paper, we observe that the key to addressing both these challenges is to leverage  composable formats and composable transformations. We propose \sys, a sparse tensor compilation abstraction that offers composable formats and composable transformations for deep learning workloads. \sys\ constructs a search space over these composable components for performance tuning. With these improvements, \sys\ obtains consistent performance speedups vs vendor libraries on GPUs for single operators: 1.20-2.34x for GNN operators, 1.05-2.98x for sparse attention operators, and 0.56-7.45x for sparse convolution operators. \sys\ also accelerates end-to-end GNNs by 1.08-1.52x for GraphSAGE training, and 4.20-40.18x for RGCN inference.

\end{abstract}




\maketitle
\section{Introduction}
\label{sec:intro}

Sparsity is becoming ubiquitous in deep learning due to the application of deep learning to graphs and the need for more efficient backbone models. Graph neural networks (GNNs) ~\cite{gcn-thomas-2017, gat-petar-2018, graphsage-hamilton-2017} have made substantial progress in modeling relations in social networks, proteins, point clouds, etc., using highly sparse matrices. Sparse transformers ~\cite{sparsetransformer-rewon-2019, longformer-iz-2020, butterfly-beidi-2021} reduce both the time and space complexity of transformers ~\cite{transformer-vaswani-2017} by making the attention mask sparse using manually designed and moderately sparse matrices. {Network Pruning ~\cite{prunning-han-2016, movement-pruning, block-prunning} prunes the network weight to sparse matrix to reduce model size, the pruned weights are moderately sparse and stored in various formats depending on the pruning algorithm.}

Existing vendor libraries, such as cuSPARSE ~\cite{cusparse}, dgSPARSE ~\cite{dgsparse}, Sputnik ~\cite{sputnik-gale-2020} and Intel MKL ~\cite{mkl}, support only a few sparse operators. As such, they fail to accelerate rapidly evolving emerging workloads such as GNNs on heterogeneous graphs ~\cite{rgcn-schlichtkrull-2017, hgat-xiao-2019, hgt-ziniu-2020} and hypergraphs ~\cite{hyper-gnn-yifan-2019}. Manually optimizing sparse operators can be difficult and tedious. Sparse matrices are stored in compressed formats, and programmers must write manual code to compress or decompress coordinates to access non-zero elements. Furthermore, the compressed sparse formats vary, and operators designed for one format cannot generalize to others. Therefore, \textit{we need a more scalable and efficient approach to developing optimized sparse operators.}

\begin{figure}[ht]
    \centering
    \includegraphics[width=0.45\textwidth]{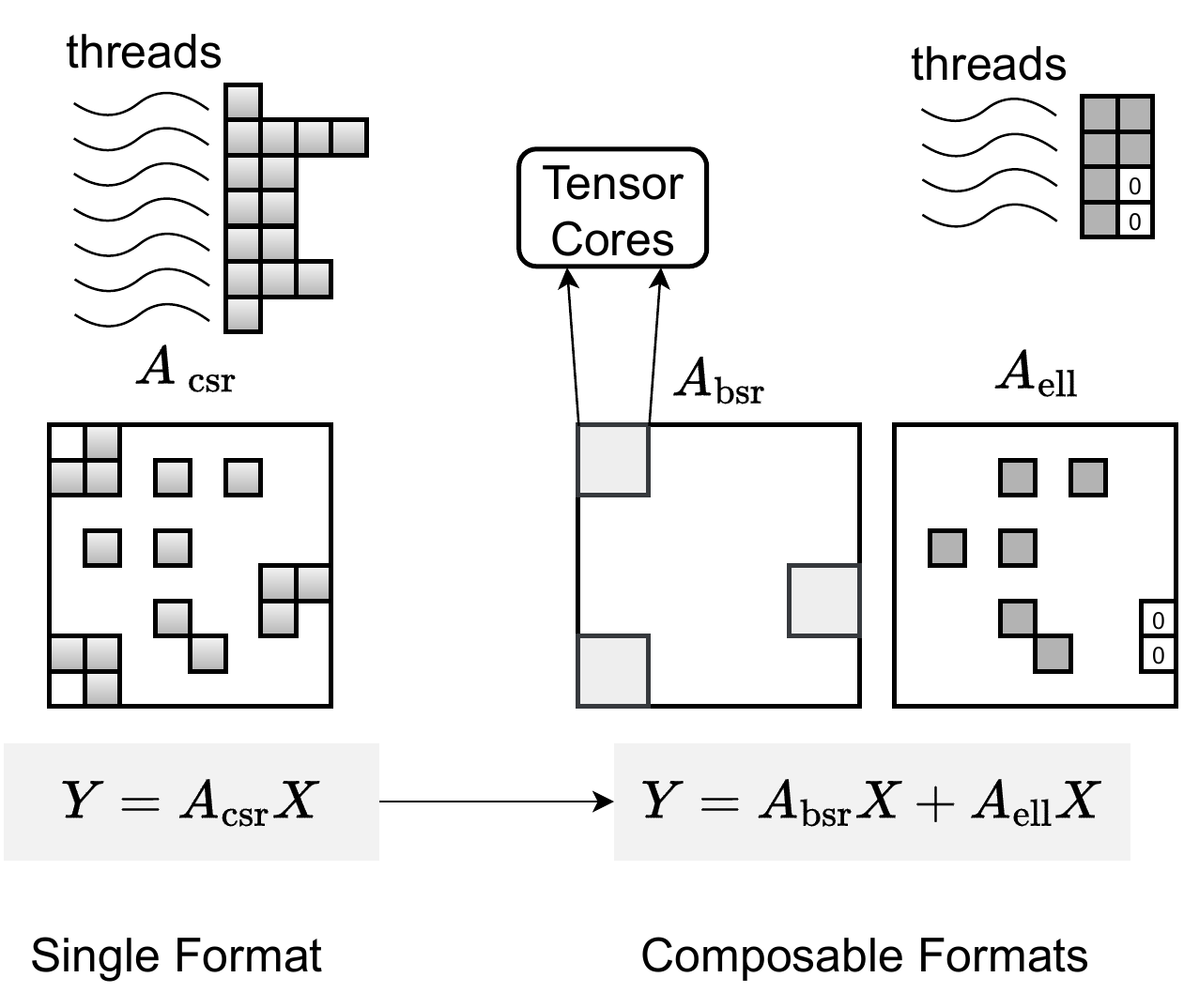}
    \caption{Format composability enables us to leverage multiple formats for different parts in sparse pattern we face in deep learning, and maximize the use of underlying hardware resources.
 }
    \label{fig:hyb-decompose.drawio.pdf}
\end{figure}

Sparse tensor compilers, such as MT1~\cite{aart-thesis} and TACO~\cite{taco-Kjolstad-2017}, greatly simplify the development of sparse operators by decoupling format specification and format-agnostic computation descriptions.
However, applying sparse
compilation to deep learning must overcome two major challenges. First, \textit{modern deep learning workloads are quite diverse,} making them hard to fit into a single sparse format pattern provided by existing solutions. Second, \textit{harware backend are evolving and becoming heterogenous}, making it hard for single-shot compilers to keep up with the latest hardware and system advances.

Our key observation is that we can resolve all challenges  by introducing two forms of  composability:

\paragraph{Format composability.} 

We propose to go beyond the single format option provided by most existing solutions to composable formats~(\autoref{fig:hyb-decompose.drawio.pdf}) that store different parts of a sparse matrix in the different formats that best fit their local patterns. The compilation process decomposes the original computations into sub-computation routines to enable efficient executions on each local pattern that better match the characteristics of the corresponding deep learning workloads.

\begin{figure*}[t]
    \centering
    \includegraphics[width=\textwidth]{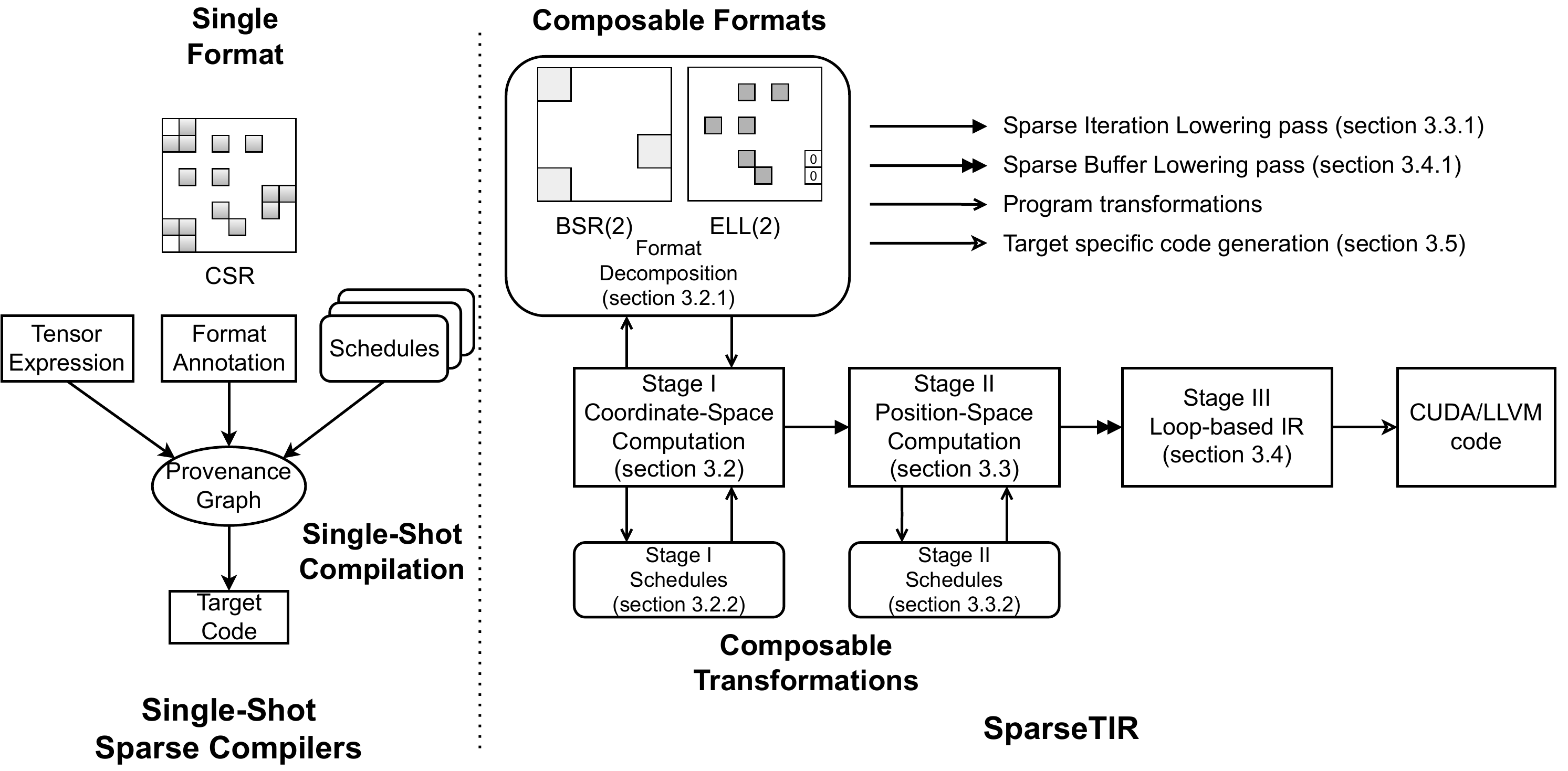}
    \caption{Single-shot sparse compilers vs \sys. The composable formats and composable transformations enable us to create optimizations that fit into broader range of deep learning workloads and leverage more advances in hardware backends.}
    \label{fig:overview}
\end{figure*}

\paragraph{Transformation composability.} 
We reconfigure the single-shot sparse tensor program compilation process into a composable set of program transformations. Additionally, we enable a design that incorporates existing loop-level abstractions in dense tensor compilers. This design lets us define our own transformations for sparse data while reusing hardware-specific optimizations (such as tensorization and GPU mapping) from existing solutions, increasing our overall efficiency to incorporate advances in hardware backends.

Combining both forms of composability, we propose \sys, an abstraction that generates efficient sparse operators for deep learning. Our contributions include the following.

\begin{itemize}
    \item We propose an intermediate representation (IR) with \textit{composable formats} and \textit{composable transformations} to accelerate sparse operators by decomposing formats and specifying schedules.
    \item We build a performance-tuning system that searches over the parameter space of possible composable formats and composable transformations.
    \item We evaluate \sys\ generated kernels on several important sparse deep learning workloads.
\end{itemize}

 \sys\ offers consistent speedup for single operators relative to vendor libraries on GPUs: 1.20-2.34x for GNN operators and 1.05-2.98x for sparse transformer operators. \sys\ also accelerates end-to-end GNNs  by 1.08-1.52x for GraphSAGE~\cite{graphsage-hamilton-2017} training and by 4.20-40.18x for RGCN ~\cite{rgcn-schlichtkrull-2017} inference, 0.56-7.45x for Sparse Convolution ~\cite{minkowskinet} operators.
\section{System Overview}
\label{sec:overview}

This section provides an overview of \sys. \autoref{fig:overview} summarizes our overall design and compares it with existing approaches. The figure's left side shows the design of most existing sparse tensor compilers ~\cite{taco-sparse-iteration-Senanayake-2020}. Their inputs are (1) tensor expressions, (2) format annotations/specifications that allow only a single format for each matrix, and (3) user-defined schedules. Schedules are applied to high-level IRs such as provenance graph, and then lowered to target device code; we refer to such compilation flow as \textit{single-shot compilation}. {These high-level IRs do not reflect low-level information such as loop structures, memory access regions, and branches. However, optimizations such as tensorization\footnote{We use this term to describe rewriting the program to use Matrix-Multiply Units such as Tensor Cores in GPU and MXU in TPU.} requires loop-level AST matching and replacement, which is not exposed in high-level IR.}  
Though tensor compilers such as Halide ~\cite{halide-ragan-2013} and TVM ~\cite{tvm-chen-2018} implement schedule primitives and code generation on multiple backends, it is difficult to re-use these infrastructures in previous sparse compilers because of the discrepancy of provenance graph and loop-level IR of existing tensor compilers.

\sys\ builds on top of these previous approaches and introduces a design that enables composable formats and composable transformations. It contains three IR stages. The first stage presents computation in coordinate space, where we describe sparse tensor computations; like in previous work, we decouple format specification and computations. Unlike a single-shot sparse compiler that accepts a single format for each sparse tensor, \sys\ lets users specify composable formats. The second stage characterizes computation in position space, where the position refers to the index of non-zero elements in the compressed sparse data structure. The concepts of ``coordinates'' and ``positions'' were first proposed in Vivienne et al. ~\cite{vivienne_springer_2020} and then used in Senanayake et al. ~\cite{taco-sparse-iteration-Senanayake-2020}.  The last stage of \sys\ is a loop-level IR in existing tensor compilers, such as TVM ~\cite{tvm-chen-2018}, AKG ~\cite{akg} and the affine dialect in MLIR ~\cite{composable-mlir-nicolas-2022}. We design two passes on the IR, namely, sparse iteration lowering and sparse buffer lowering, to transform code from stage I to stage II and stage II to stage III, respectively.

Instead of single-shot compilation, all schedules in \sys\ are performed as composable program transformations (which do not change the stage of the IR) on the IR instantly.
The composable design lets user transform the IR step-by-step and stage-by-stage. To manipulate the coordinate space computation in stage I IR, we can define new schedules as composable transformations applied to the stage I (i.e., stage I schedules). For stages  compatible with target loop-level IR, we can apply schedules defined for backend tensor compilers (i.e., stage II/III schedules).
Notably, format decomposition can also be formulated as a program transformation at stage I (see \S \ref{sec:format-decomposition}). 

\sys\ constructs a joint search space of composable formats and composable transformations for performance tuning of sparse operators. Users can customize the parameterized search space by specifying format and schedule templates based on their domain-specific knowledge about the operator and sparse tensor characteristics. When the sparse structure is present at compile-time, we can search for the best formats and schedules that achieve optimal runtime performance in advance. Though the compilation might take some time due to the large search space, the overhead can be amortized because the compiled operator will be re-used many times during training or inference for a fixed sparse structure (as is typical in deep learning).

The rest of the paper is organized as follows.
We introduce the \sys\ design of each stage and compiler passes in Section \ref{sec:compilation}. In Section \ref{sec:eval} we evaluate our system in real world sparse deep learning workloads. Section \ref{sec:related-work} positions \sys\ relative to related work. Finally, we discuss future work in Section \ref{sec:future-work} and conclude our work in Section \ref{sec:conclusion}.

\section{Our Approach}
\label{sec:compilation}

In this section, we introduce the language constructs in \sys, then describe each compilation stage and transformations in the order they appeared in the flow.

\subsection{Language Constructs}
\label{sec:lang-construct}

\begin{figure}[ht]
    \centering
    \includegraphics[width=0.45\textwidth]{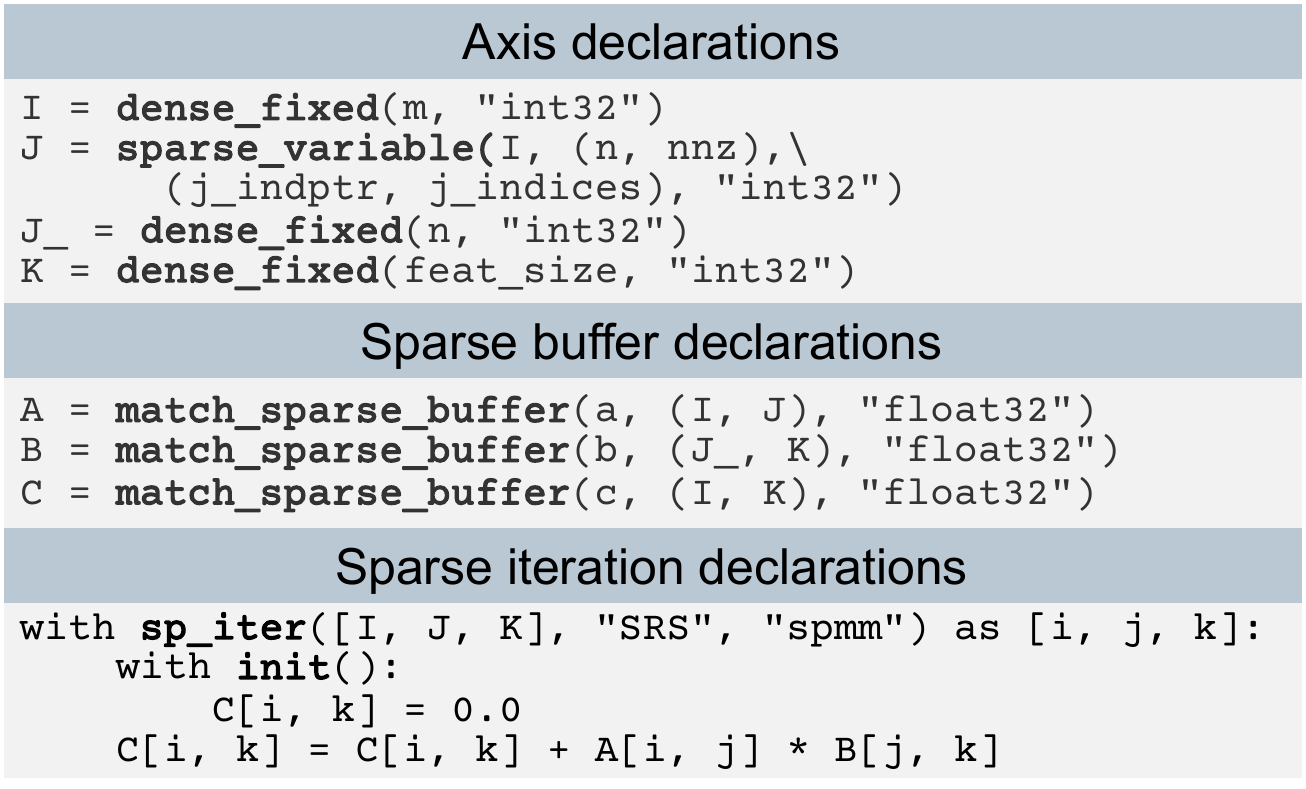}
    \caption{Language constructs in the SpMM operator. Users specify axis dependencies and metadata to create axes. The \lstinline{match\_sparse\_buffer} defines sparse buffers and binds them to pointers to their value, and \lstinline{sp\_iter} creates a sparse iteration structure, where ``S'' and ``R'' indicate whether the iterator is for spatial or reduction purposes, ``spmm'' is the name of the sparse iteration as a reference for scheduling.}
    \label{fig:lang-construct}
\end{figure}

The \sys\ language has three major components: axes, sparse buffers and sparse iterations.

\paragraph{Axes.} An \textit{axis} is a data structure that defines sparse iteration spaces, which generalize the idea of abstraction levels in previous work ~\cite{format-abstraction-chou-2018}. Each axis in \sys\ has two orthogonal attributes, dense/sparse and fixed/variable, denoting whether the index of non-zero elements in the  axis is contiguous or not and whether the number of non-zero elements in the axis is fixed or not. Variable axes are associated with a \lstinline{indptr} {(short for ``index pointer'')} field that  points to the address of the indices pointer array; sparse  axes are associated with an \lstinline{indices} field that  points to the address of the indices array. Each axis has a parent field that directs to the axis it depends on; a  dense-fixed axis has no dependency, and its parent field is always set to none. Axis metadata includes its indices' data type, maximum length, number of accumulated non-zeros in this dimension (if variable), and number of non-zeros per row in this dimension (if fixed).

\begin{figure}[ht]
    \centering
    \includegraphics[width=0.38\textwidth]{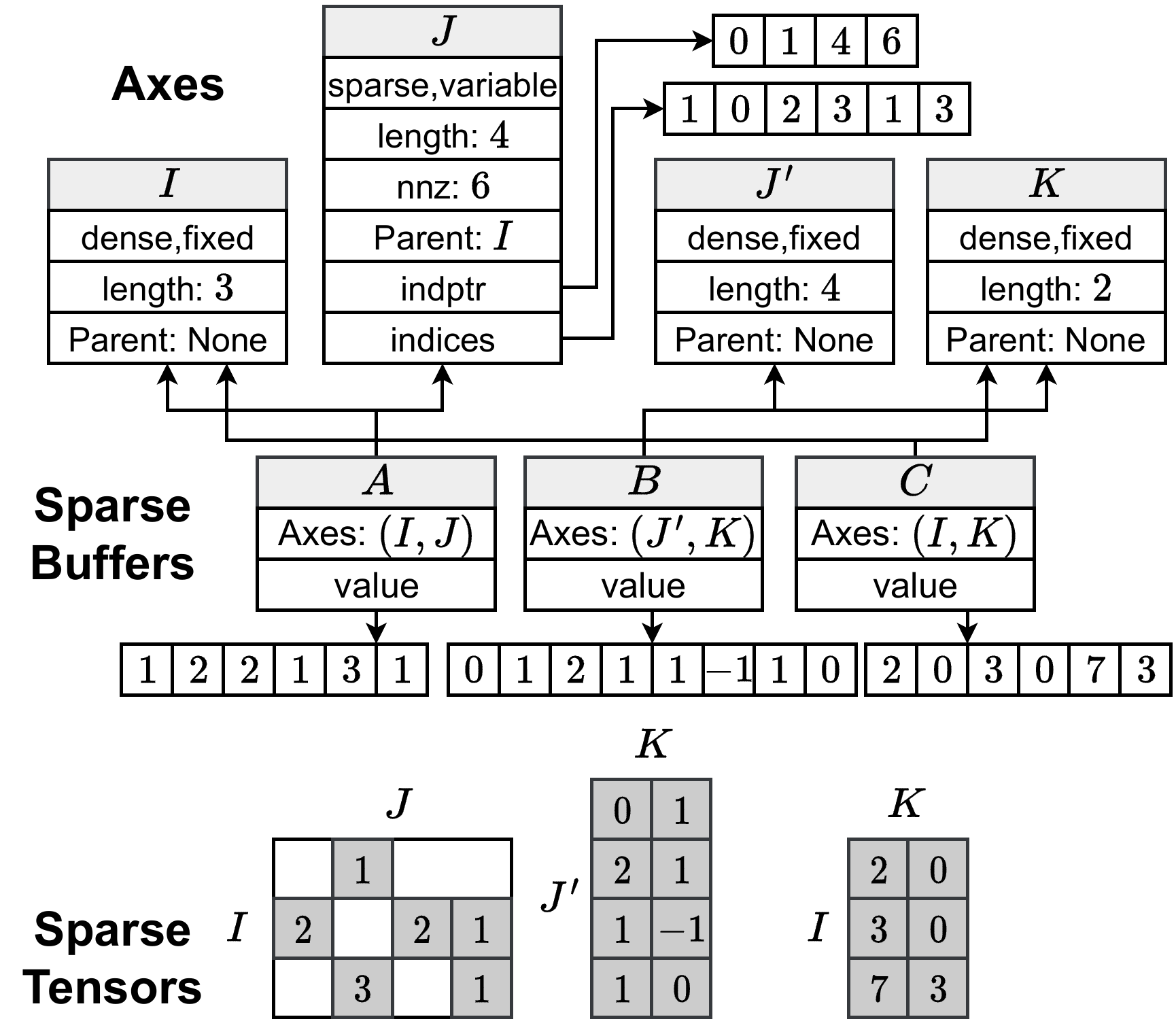}
    \caption{Internal storage of axes and sparse buffers in SpMM: $C_{ik} = A_{ij}B_{jk} $. Sparse buffers store their axes' composition and pointers to their value; axes store dense/sparse and fixed/variable attributes, metadata, their dependent axes, and pointers to indices and indptr arrays.}
    \label{fig:sparse-storage}
\end{figure}

\paragraph{Sparse buffers.} A sparse buffer is \sys's data structure for a sparse matrix. We use defined axes to compose the format specification of sparse matrices. We split sparse structure-related auxiliary data and values: axes store auxiliary data, and sparse buffers store only values. Such design lets Two sparse buffers can re-use auxiliary data if they share the sparse layout. Figure \ref{fig:sparse-storage} shows the decoupled storage of sparse buffers/axes in the SpMM (Sparse-Dense Matrix Multiplication) operator. The composition of axes is expressive to describe various sparse formats, including Compressed Sparse Row/Column (CSR/CSC) format~\cite{duff1986direct}, Block Compressed Sparse Row (BSR) format ~\cite{saad1990sparskit}, Diagonal Format (DIA) ~\cite{saad1989krylov}, ELLPACK (ELL) format ~\cite{Duff1987,oppe1987itpack,ellpack}, Ragged Tensor ~\cite{ragged-tensor}, Compressed Sparse Fiber (CSF) ~\cite{csf} etc, please refer to Duff et al. ~\cite{duff1986direct} for an overview of sparse formats.

\paragraph{Sparse iterations.} Sparse iterations generates iterators  over the space composed of a sparse axes array and a body containing statements describing tensor computations and orchestrating data movements. Notably, unlike TACO ~\cite{taco-Kjolstad-2017} 
which only allows the iterator variable to be used as an index to access sparse data structures (e.g. \lstinline{A[i, j]} where \lstinline{i} and \lstinline{j} are iterator variables), \sys\ supports any expressions, including affine indices (e.g. \lstinline{A[i * m + j, k]}) and integer values loaded from another buffer (e.g. \lstinline{B[eid[i], j * n + k]}). This enhances the capabilities of the \sys, allowing for more complex operations such as convolution. \sys\ enables multiple sparse iterations within a single program and even allows for nested sparse iterations within the body of another iteration, enabling branching and decomposing computation.

Figure \ref{fig:lang-construct} shows how to define these constructs in \sys\ for the SpMM operator.\footnote{The \sys\ has round-trip compatibility with Python, and this paper presents only its Python form.}
In \sys, axes are used to construct both sparse buffers and sparse iterations. This design lets us iterate over a sparse iteration space that is not bound to any sparse buffers. 

\subsection{Stage I: Coordinate Space Computation}

In stage I \sys\ defines sparse computations inside sparse iterations, where we iterate over non-zero elements and access sparse buffers in the coordinate space. At this stage, we can define program transformations,  such as format decomposition and sparse iteration fusion, that manipulate only the three constructs of the \sys\ .

\subsubsection{Format Decomposition}
\label{sec:format-decomposition}

\begin{figure}[ht]
    \centering
    \includegraphics[width=0.45\textwidth]{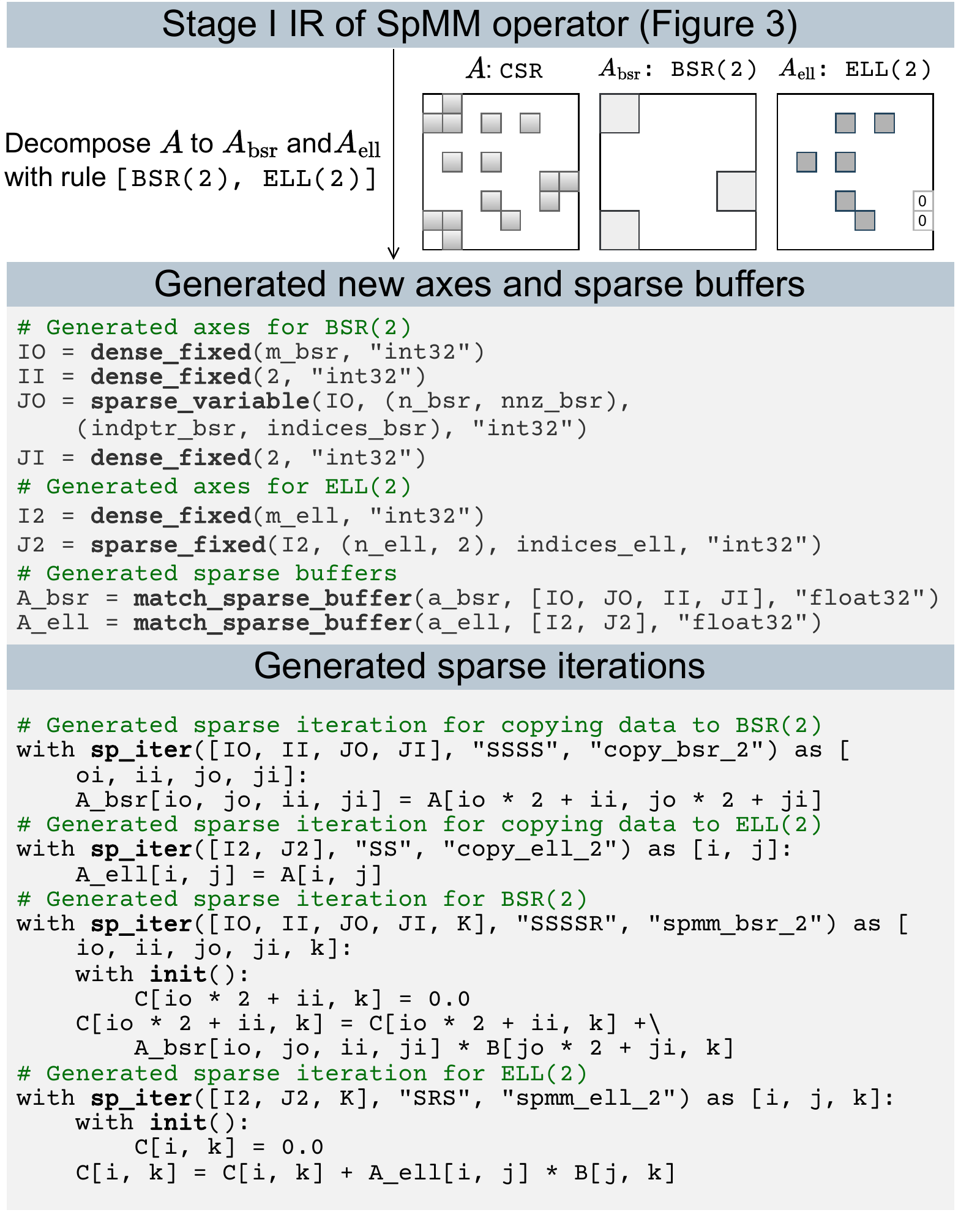}
    \caption{Format decomposition for SpMM Stage I IR in Figure \ref{fig:lang-construct}. New axes and sparse buffers are created for decomposed formats BSR and ELL. New sparse iterations are generated to copy data from original to new formats and for computations on these new formats.}
    \label{fig:format-decomposition}
\end{figure}

\textit{Format decomposition} is a transformation that decomposes  computations for composable formats (introduced in Section \ref{sec:intro}).
The transformation accepts a list of format descriptions and rewrites the IR according to these formats. Figure \ref{fig:format-decomposition} shows  the generated IR for the Sparse Matrix-Matrix multiplication (SpMM) operation after decomposing the computation in the CSR format to a computation in the BSR format, with block size $2$ and an ELL format with $2$ non-zero columns per row. In addition to SpMM computations on the new formats, another two sparse iterations that copy data from original to new formats are generated, as well. {When the sparse matrix to decompose is stationary, we can perform data copying at pre-processing step to avoid the overhead of run-time format conversion.}

{The information used to create new sparse buffers: \lstinline{indptr_bsr}, \lstinline{indices_bsr} and \lstinline{indices_ell} need to be pre-computed and specified by user as input arguments. Each format decomposition rule in \sys\ needs to be registered as a 
function $F: (x, i)\rightarrow (x', i')$, where $x,i$ refers to original \sys\ program and indices/index pointer information, and $x',i'$ are transformed ones. Figure \ref{fig:format-decomposition} describes the IR transformation from $x$ to $x'$, and the conversion between $i$ to $i'$ need to be implemented by user manually. We have wrapped all format decomposition rules used in this paper as standard APIs, for new composable formats, user can use existing sparse libraries such as Scipy ~\cite{scipy} to ease the implementation of indices inference. \sys\ leaves the flexibility of integrating with existing systems such as Chou et al. ~\cite{taco-conversion-2020} for automatic indices inference.}

\subsubsection{Stage I Schedules}
\label{sec:stage-i-schedules}
We define two schedule primitives at stage I,  \lstinline{sparse_reorder} and \lstinline{sparse_fuse}.

\begin{figure}[ht]
    \centering
    \includegraphics[width=0.45\textwidth]{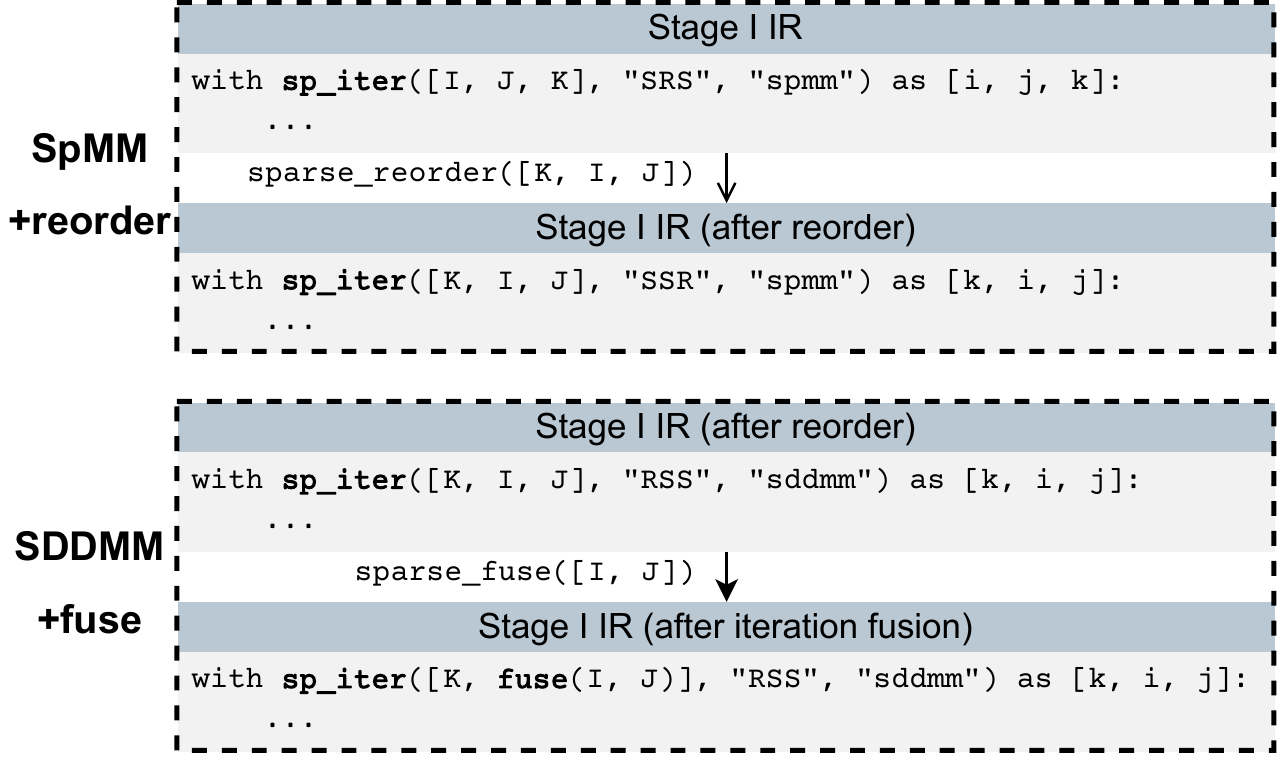}
    \caption{Stage I schedules sequentially applied to stage I IR.}
    \label{fig:stage-i-schedules}
\end{figure}

\paragraph{Sparse reorder.}
\label{sec:sparse-reorder}
The order of sparse axes in the sparse iteration  influences the order of generated loops in stage II. This primitive enables manipulation of the order of sparse axes.

\paragraph{Sparse fuse.}
\label{sec:sparse-fuse}
This schedule primitive fuses several iterators in a given sparse iteration into one. It is helpful when we want a single loop rather two nested loops that iterate over all non-zero elements, such as in the SDDMM~\cite{sddmm-nisa-2018}.

Figure \ref{fig:stage-i-schedules} shows how stage I schedules transform the IR.

\subsection{Stage II: Position Space Computation}

The State II IR in \sys\ introduces loop structures and removes the sparse iteration constructs and restructuring them as nested loops. Unlike in stage I where we access sparse buffers in \textit{coordinate space}, in stage II access sparse buffers in \textit{position space}, with the ``position'' referring to an element's non-zero index. The difference between coordinate and position applies to ``sparse'' dimensions: if the coordinate of the first 4 non-zero elements in a sparse row $A$ is $\{1, 3, 9, 10\}$, the position of $9$ is $2$ (assuming the index is 0-based), and we use $A[9]$ to access the element in coordinate space and $A[2]$ to access the element in position space. Our stage II IR extends TensorIR~\cite{tensorir} and treats sparse buffer as first-class citizens.

\subsubsection{Sparse Iteration Lowering.}
\label{sec:sparse-iteration-lowering}

This pass transforms stage I IR to stage II IR. It consists of the following 4 steps.

\begin{figure}[ht]
    \centering
    \includegraphics[width=0.45\textwidth]{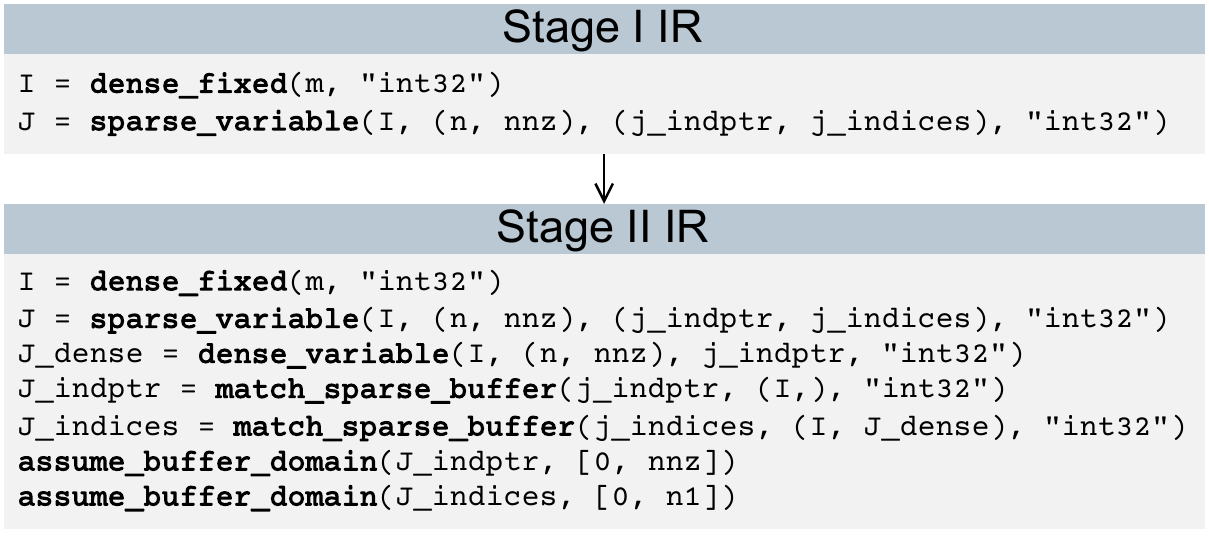}
    \caption{Example of auxiliary buffer materialization. Sparse buffers storing auxiliary information are created.}
    \label{fig:aux-buf-materialize}
\end{figure}

\paragraph{Step 1: Auxiliary buffer materialization.} 
Pointers to the indices pointer array and indices array are specified as arguments when creating axes. In stage II we need to declare these auxiliary buffers explicitly to access their value when determining loop range and translating coordinates. Figure \ref{fig:aux-buf-materialize} shows how the materialization works. In addition to auxiliary buffers, we also create hints that indicate the domain of buffer values; these are used for integer set analysis in stage II when performing schedules.

\paragraph{Step 2: Nested loop generation.}

\begin{figure}[ht]
    \centering
    \includegraphics[width=0.45\textwidth]{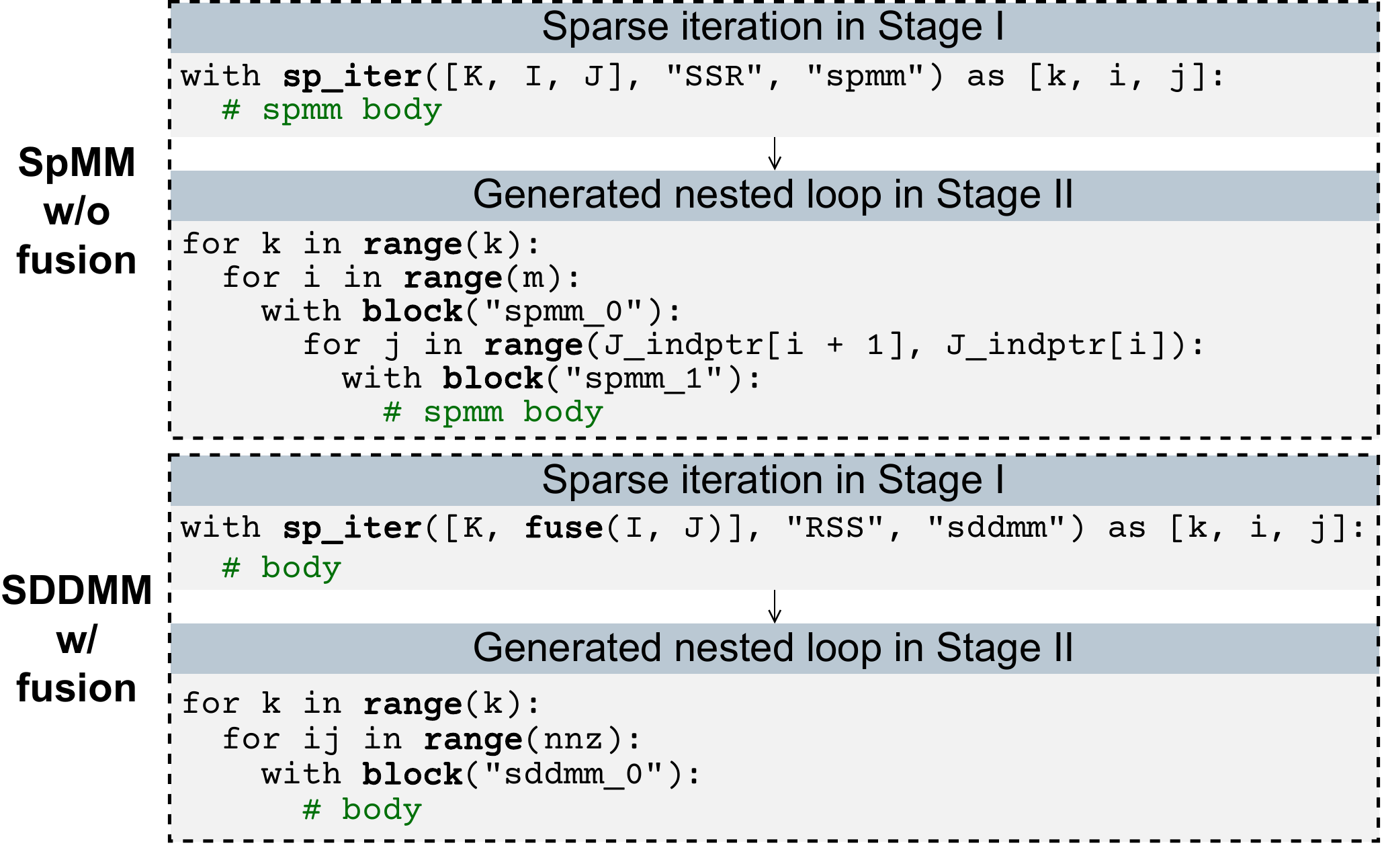}
    \caption{Nested loop generation in sparse iteration lowering. Without fusion, we emit one loop per axis in the sparse iteration; With fusion of $i$ and $j$, we only emit one loop $ij$ over the fused iteration space.}
    \label{fig:nested-loop-generation}
\end{figure}

This step restructures sparse iterations in stage I as nested loops in stage II: we emit one loop per axis in the sparse iteration. The generated loops start from $0$, and the extent is determined by whether the axis is fixed or variable. They are separated by TensorIR's \lstinline{block} constructs, which establish boundaries to prevent cross-block loop reordering. Additionally, We add a \lstinline{block} inside the innermost generated loop and place the body of original sparse iterations inside of it.
Figure \ref{fig:nested-loop-generation} shows the emitted nested loop structures of different sparse iterations. In the first case, the loops $I$ and $J$ cannot not be reordered in stage II because they are separated by a block; in the second case, we fuse $I, J$ and emit only one loop (\lstinline{ij}).

\begin{figure}[ht]
    \centering
    \includegraphics[width=0.45\textwidth]{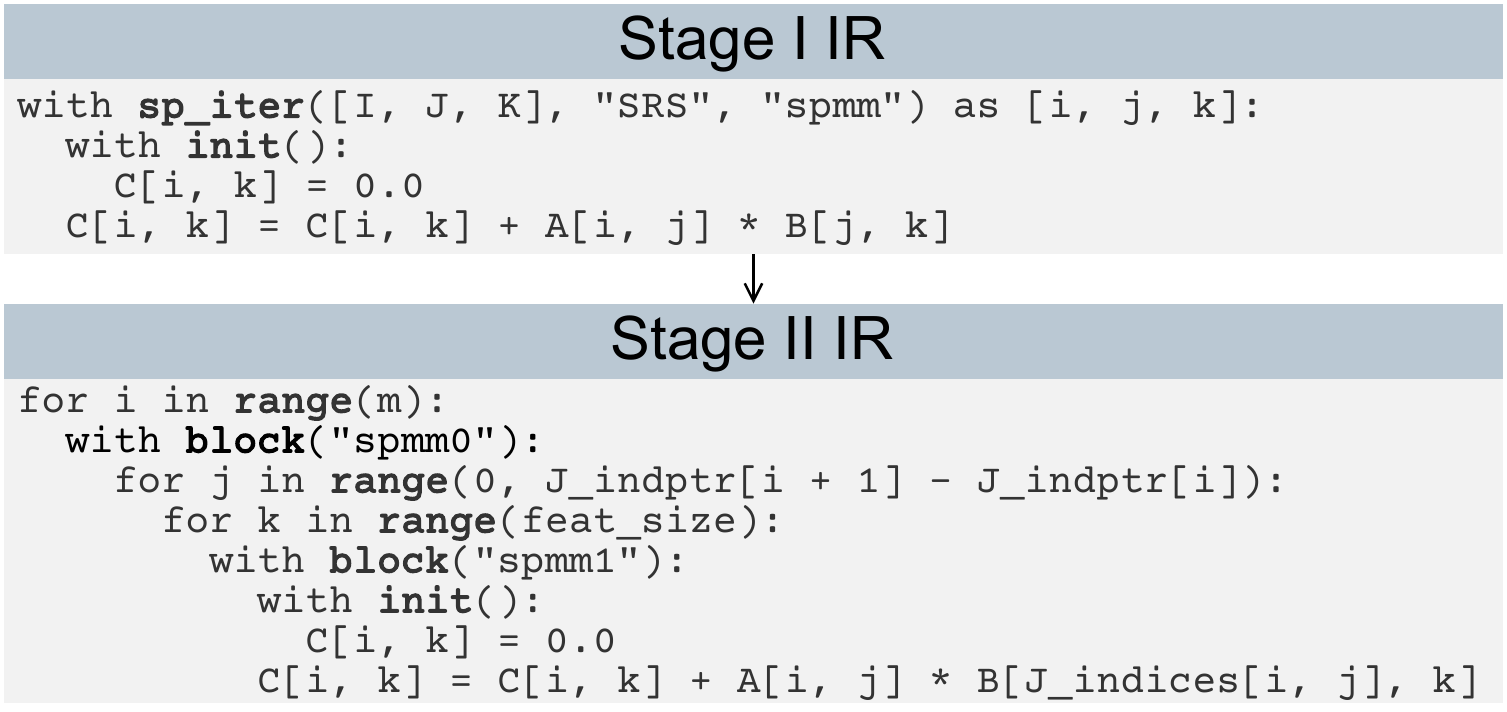}
    \caption{Translation from coordinate space to position space for SpMM operator.}
    \label{fig:coord-trans}
\end{figure}

\paragraph{Step 3: Coordinate translation.}
This step rewrites the indices used to access sparse buffers from coordinate space to non-zero position space to bridge the semantic gap between stages I and  II. See Figure \ref{fig:coord-trans} for an example. Suppose $\{ \mathbf{A}^{(\textrm{iter})}_i \}_{i = 1}^M$ is the array of axes
used in sparse iterations, $\{ \mathbf{v}^{(c)}_i \}_{i = 1}^M$ is the array
of iterator variables in coordinate space (before translation) and $\{
\mathbf{v}^{(p)}_i \}_{i = 1}^M$ is the array of loop variables in position
space (after translation). For a sparse buffer access to be translated,
suppose the buffer is composed of axes $\{ \mathbf{A}^{(\textrm{buffer})}_j
\}_{j = 1}^N$, and the indices can be viewed as an array of functions $\{
\mathbf{I}^{(\textrm{coord})}_j \}_{j = 1}^N$ that maps iterator varibles to
indices (for buffer access $B [x + y, z]$ within the sparse iteration where
$\mathbf{v}^{(c)} = \{ x, y, z \}$, its indices functions
$\mathbf{I}^{(\textrm{coord})}$ should be $\{ (x, y, z) \mapsto x + y, (x, y,
z) \mapsto z \}$). The coordinate translation can be formulated as an iterative algorithm:

\begin{equation}
    \mathbf{p}_j \triangleq f^{- 1} (\mathbf{A}^{(\textrm{buffer})}, j, \{
        \mathbf{p} \}_1^{j - 1}, \mathbf{I}^{(\textrm{coord})} (\mathbf{c}_1, \ldots,
        \mathbf{c}_M))
\end{equation}

\noindent where $\mathbf{c}$ refers the coordinate array corresponding to $\mathbf{v}^{(p)}$ after translation from position space, $f$ and $f^{(-1)}$ are decompress (position to coordinate) and compress (coordinate to position) functions:

\begin{equation}
    \mathbf{c}_i \triangleq f (\mathbf{A}^{(\textrm{iter})}, \{\mathbf{c}\}_1^{i-1}, \mathbf{v}^{(p)}_i)
\end{equation}

\begin{equation}
    f (\mathbf{A}, i, \mathbf{c}, x)
\triangleq 
\left\{
    \begin{array}{ll}
      x & \mathbf{A}_i \mathbf{} \text{: D(ense)}\\
      \mathbf{A}_i\textrm{\_indices} [\mathbf{c}[\textrm{anc} (\mathbf{A},
      i)], x] & \mathbf{A}_i \mathbf{} \text{: S(parse)}
    \end{array} \right.
\end{equation}

\begin{equation}
    f^{(- 1)} (\mathbf{A}, j, \mathbf{p}, x) 
    \triangleq 
    \left\{ \begin{array}{ll}
        x & \mathbf{A}_j \text{: D}\\
        \textrm{find}(\mathbf{A}_j\textrm{\_indices}[\mathbf{p}[\textrm{anc}
        (\mathbf{A}, j)],:], x) & \mathbf{A}_j \text{: S} \label{eq:bsearch}
      \end{array} \right.
\end{equation}

\noindent The ``find'' function in the later case of equation \ref{eq:bsearch} refers to searching a given value in sorted array, \sys\ emits a binary search \lstinline{block} to search for the index of $x$ in sorted indices array. The ``anc'' function collects the indices of ancestor(including self) axes of $\mathbf{A}_i$ from its root in axes dependency tree, and $\mathbf{p}[\textrm{anc}(\mathbf{A}, j)]$ gathers values from $\mathbf{p}$ by ancestors' indices:

\begin{equation}
    \textrm{anc} (\mathbf{A}, i) \triangleq \left\{
        \begin{array}{ll}
          \lbrack i \rbrack & \mathbf{A}_i  \text{ is root}\\
          \lbrack \textrm{anc} (\mathbf{A}, j) : i \rbrack &
          \mathbf{A}_j = \textrm{parent} (\mathbf{A}_i)
        \end{array} \right.
\end{equation}

\paragraph{Read/Write Region Analysis}

The buffer read/write region information is necessary for TensorIR's \lstinline{block} construct. We perform a buffer region analysis pass to collect buffer access information and takes the union of all read/write regions accessed inside each block and annotate them as block attributes.

\subsubsection{Stage II Schedules}
\label{sec:stage-ii-schedules}
The stage II schedules are responsible for manipulating loops (\lstinline{fuse}/\lstinline{reorder}/\lstinline{split}), moving data across the memory hierarchy (\lstinline{cache_read}/\lstinline{cache_write}), binding loops to physical/logical threads to parallelize them, and using vector/tensor instructions in hardware (\lstinline{vectorize}/\lstinline{tensorize}). As a dialect of TensorIR, we fully support TVM schedule primitives \footnote{\url{https://tvm.apache.org/docs/reference/api/python/tir.html\#tvm.tir.Schedule}} at stage II.

\subsection{Stage III: Loop-Level IR}
Stage III removes all  \sys\ constructs. It keeps only the nested loop structures whose body includes statements that operate on flattened buffers. This stage should be compatible with loop-level IR in existing tensor compilers. We select TensorIR ~\cite{tensorir} in Apache TVM ~\cite{tvm-chen-2018} as stage III IR to make efficient use of NVIDIA's Tensor Cores, as it fully supports tensorization.

\subsubsection{Sparse Buffer Lowering}
\label{sec:sparse-buffer-lowering}
\begin{figure}[ht]
    \centering
    \includegraphics[width=0.45\textwidth]{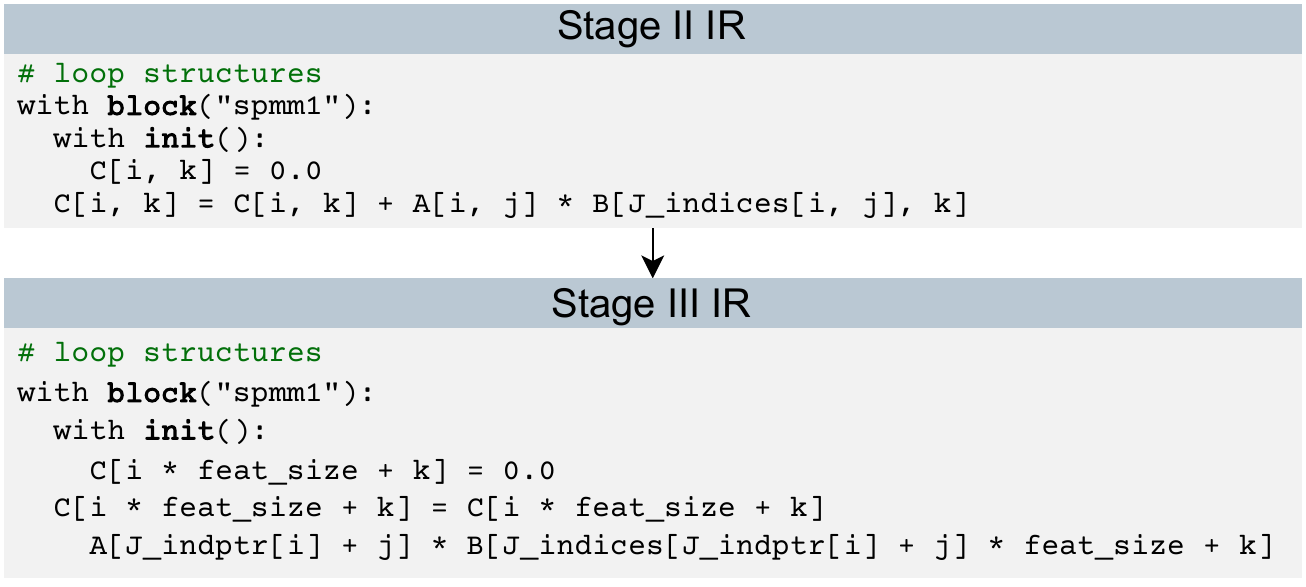}
    \caption{Sparse buffer lowering: sparse constructs are totally removed, and memory accesses are flattened to 1-dimension.}
    \label{fig:sparse-buffer-lowering}
\end{figure}

Sparse buffer lowering removes all axes, flattens all multi-dimensional sparse buffers to 1-dimension, and rewrites memory access to these buffers.
Suppose the original sparse buffer $A$ is composed of axes $\{\mathbf{A}_i\}_{i=1}^{n}$. For memory access $A[x_1, ..., x_n]$, the overall offset after flattening is computed by:

\begin{equation}
\sum_{i=1}^{n} \textrm{is\_leaf}(\mathbf{A}_i) \times \textrm{offset}(i) \times \textrm{stride}(i + 1)
\label{eq:sp-buffer-lower}, 
\end{equation}
\noindent where is\_leaf$(\mathbf{A}_i)$ means that  if axis $\mathbf{A}_i$ has no dependence in $\{\mathbf{A}_j\}_{j=i+1}^{n}$, offset and stride are defined as:
\begin{equation}
\textrm{offset}(i) \triangleq 
\begin{cases}
    x_i & \textrm{is\_root}(\mathbf{A}_i) \\
    \mathbf{A}_i\textrm{\_indptr} [\textrm{offset}(j)] + x_i & \mathbf{A}_j=\textrm{parent}(\mathbf{A}_i)  
\end{cases}    
\end{equation}
\begin{equation}
\textrm{stride}(i) \triangleq
\begin{cases}
    1 & i > n \\
    \textrm{nnz}(\textrm{Tree}(\mathbf{A}_{i})) \times \textrm{stride}(i+1) & \textrm{is\_root}(\mathbf{A}_{i}) \\
    \textrm{stride}(i+1) & \textrm{otherwise},
\end{cases}  
\end{equation}
where $\textrm{nnz}(\textrm{Tree}(\mathbf{A}_i))$ refers to the number of non-zero elements of the sparse iteration space composed by the tree with $\mathbf{A}_i$ as its root. Figure \ref{fig:sparse-buffer-lowering} shows an example of sparse buffer lowering: sparse buffers $A$, $B$, $C$ are flattened. The buffer access $A[i, j]$ is translated to $A[\textrm{J\_indptr}[i]+j]$ by equation \ref{eq:sp-buffer-lower}.

\subsection{Target-Specific Code Generation}
\label{sec:target-specific-codegen}

\sys\ re-uses the backend provided by existing tensor compilers for target-specific code generation. \sys\ emits multiple CUDA kernels for composable formats, which incur extra kernel-launching overhead on the GPU. We insert a horizontal fusion ~\cite{horizontal-fuse, cora-pratik-2021} pass to the TVM backend to reduce this overhead.

\section{Evaluation}
\label{sec:eval}

We now study how composable formats and composable transformations help optimize sparse deep learning workloads in both single-operator and end-to-end settings. In summary, compared to vendor libraries, \sys\ obtains a 1.20-2.34x speedup on GNN operators and a 1.05-2.98x speedup on sparse attention operators. When used in an end-to-end setting, \sys\ obtains a 1.08-1.52x speedup on end-to-end GraphSAGE training and a 4.20-40.18x speedup on end-to-end RGCN inference, 0.56-7.44x on Sparse Convolution operators.

\subsection{Experiment Setup}

\paragraph{Environment.}
\label{sec:environment}
We evaluate all experiments under two different GPU environments: NVIDIA RTX 3070 and NVIDIA Tesla V100.

\paragraph{Baselines.}
\label{sec:baselines}

cuSPARSE~\cite{cusparse} is NVIDIA's official library for sparse tensor algebra, which includes high-performance implementation of common sparse operators. dgSPARSE~\cite{dgsparse} is a collection of state-of-the-art sparse kernel implementations for GNNs, which includes GE-SpMM~\cite{ge-spmm-huang-2020}, DA-SpMM~\cite{spmm-heuristic-adaptability-2022} and PRedS~\cite{sddmm-yu-2021}. PyG~\cite{pyg-fey-2019} and DGL~\cite{dgl-wang-2019} are two open-source frameworks that support GNN training and inference. {Sputnik~\cite{sputnik-gale-2020} is a library for sparsity in Deep Learning. Neither dgSPARSE nor Sputnik uses Tensor Cores.} TACO~\cite{taco-Kjolstad-2017} is an open-source sparse tensor compiler. Triton~\cite{triton-tillet-2019} is a tiling-based IR for programming neural networks, and we use its block sparse operator implementation. {TorchSparse ~\cite{torchsparse-tang-2022} is a library for point cloud processing, with state-of-the-art sparse convolution implementation.}

For SpMM, we select the TACO-generated operator, cuSPARSE 11.7, and dgSPARSE 0.1 as baselines. For SDDMM, we select the TACO-generated operator, cuSPARSE, dgSPARSE and DGL 0.9.1's implementation as baselines. The DGL's SDDMM implementation uses the optimizations proposed in FeatGraph ~\cite{featgraph-hu-2020}. For end-to-end GNN training, we compare a GraphSAGE model written in PyTorch 1.12 ~\cite{pytorch} that integrates a \sys-tuned kernel with DGL. For RGCN, we select the Graphiler ~\cite{graphiler-xie-2022}, DGL 0.9.1 and PyG 2.2.0 implementations as our baseline.\footnote{Both DGL and PyG provide several different official implementations of RGCN; we select the best performing among them.}  For sparse transformers, we select Triton\footnote{Main branch until commit \hyperlink{https://github.com/openai/triton/commit/0e8590f1c9d3c3301187d2b450714f8ca8b4eeda}{0e8590}}'s block-sparse kernel as our baseline. {For sparse convolution, we select TorchSparse \footnote{Main branch until commit \href{https://github.com/yzh119/torchsparse/commit/2caf0846b8be3df24e851a11b6c580db81f5f4ff}{2caf084}} for comparison.} The computation results of all \sys\ generated kernels have been compared with existing frameworks/libraries to confirm numerical accuracy.

\subsection{Graph Neural Networks}
In this section, we evaluate the performance of \sys\ on GNN workloads. SpMM and SDDMM ~\cite{sddmm-nisa-2018} are two of the most generic operators in GNNs. Table \ref{tab:datasets} describes the characteristics of graphs used in our evaluation; on the table, {\%padding refers to the ratio of padded zero elements after we transform the original sparse matrix to composable formats.}

\begin{table}[t]
\small
\centering
\begin{tabular}{llll}
\toprule
Graph         & \#nodes   & \#edges & \%padding \\
\midrule
cora ~\cite{pubmed}         & 2,708     & 10,556 & 15.9 \\
citeseer ~\cite{pubmed}     & 3,327     & 9,228 & 13.0 \\
pubmed ~\cite{pubmed}       & 19,717    & 88,651 & 23.1 \\
ppi ~\cite{graphsage-hamilton-2017}          & 44,906    & 1,271,274 & 22.9 \\
ogbn-arxiv~\cite{hu2020ogb}    & 169,343   & 1,166,243 & 17.5 \\
ogbn-proteins~\cite{hu2020ogb} & 132,534   & 39,561,252 & 21.6 \\
reddit ~\cite{graphsage-hamilton-2017}       & 232,965   & 114,615,892 & 28.6 \\
\bottomrule
\end{tabular}
\caption{Statistics of Graphs used in GNN experiments.}
\label{tab:datasets}
\end{table}

\subsubsection{SpMM}
\label{sec:spmm-eval}

SpMM is the most generic sparse operator in deep learning, which can be formulated as:

$$ Y_{i,k} = \sum_{j=1}^{n}A_{i,j} X_{j,k}, $$
\noindent where $A$ is a sparse matrix and $X, Y$ are dense matrices. A high-performing SpMM kernel on a GPU requires efficient memory access patterns and load balancing~\cite{merge-spmm-carl-2018}. Runtime load balancing, well studied in SpMM acceleration literature, always incurs runtime overhead. The composable format and composable transformation can help generate kernels that achieve compile-time load balancing and better cache utilization.

\begin{figure}[ht]
    \centering
    \includegraphics[width=0.4\textwidth]{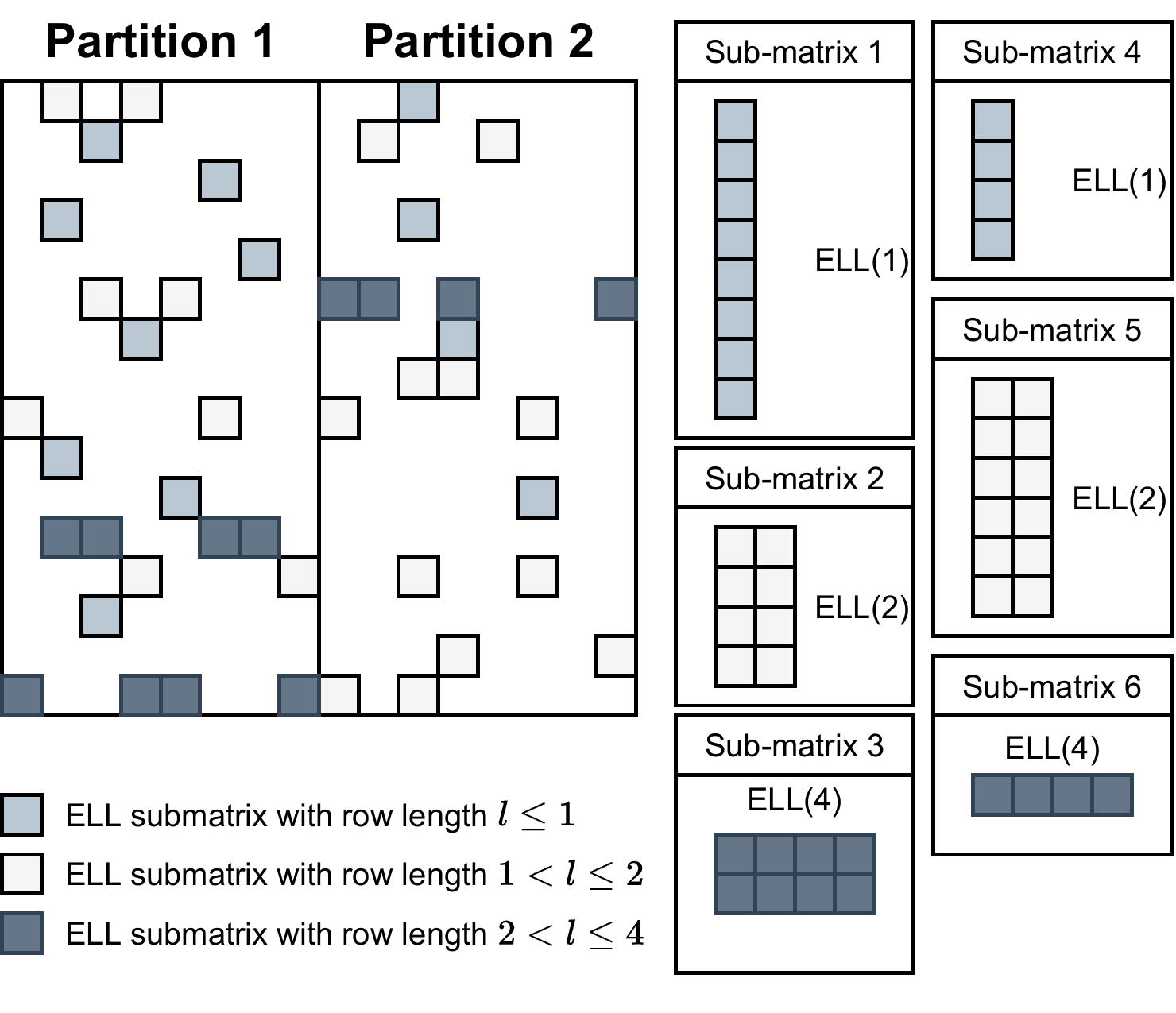}
    \caption{Example of $hyb(2, 2)$: the original matrix is decomposed to $6$ ELLPACK sub-matrices; elements in partition 1 are stored in sub-matrix 1-3, and elements in partition 2 are stored in sub-matrices 4-6.}
    \label{fig:spmm-hyb-format}
\end{figure}

We design a parameterized composable format $hyb(c, k)$ for sparse matrix $A$ with two parameters $c$ and $k$. We partition columns of the sparse matrix by the given factor $c$, so that each column partition has width $w$. For each column partition, we collect the rows with length $l$ that satisfy $2^{i-1} < l \leq 2^{i}$ to bucket $i$, and we pad the length of these rows to $2^{i}$; each bucket then forms a sub-matrix with the ELL format. Figure \ref{fig:spmm-hyb-format} shows a special case, $\textrm{hyb}(2, 2)$.

For bucket $i$ of each column partition, we group each $2^{k-i}$ rows and map them to a unique thread block in GPUs. The number of non-zero elements in $A$ that are processed by each thread block is $2^k$, which is implemented with TVM's \lstinline{split} and \lstinline{bind} primitives. We use the schedule proposed in GE-SpMM ~\cite{ge-spmm-huang-2020} for each sub-matrix for the remaining dimensions. The column partition in our design is intended to improve cache locality; when processing column partition $j$, only $B[j w: (j+1)w]$ would be accessed for $B$. Featgraph ~\cite{featgraph-hu-2020} proposes to apply column partitions for SpMM on CPUs; however, it does not extend the idea to GPUs. Our bucketing technique was designed to achieve compile-time load balancing. 
In practice, we searches for the best $c$ over $\{1, 2, 4, 8, 16\}$ and let $k = \left\lceil \log_{2} \frac{nnz}{n} \right\rceil$, which generally works well.

\begin{figure}[ht]
    \centering
    \includegraphics[width=0.45\textwidth]{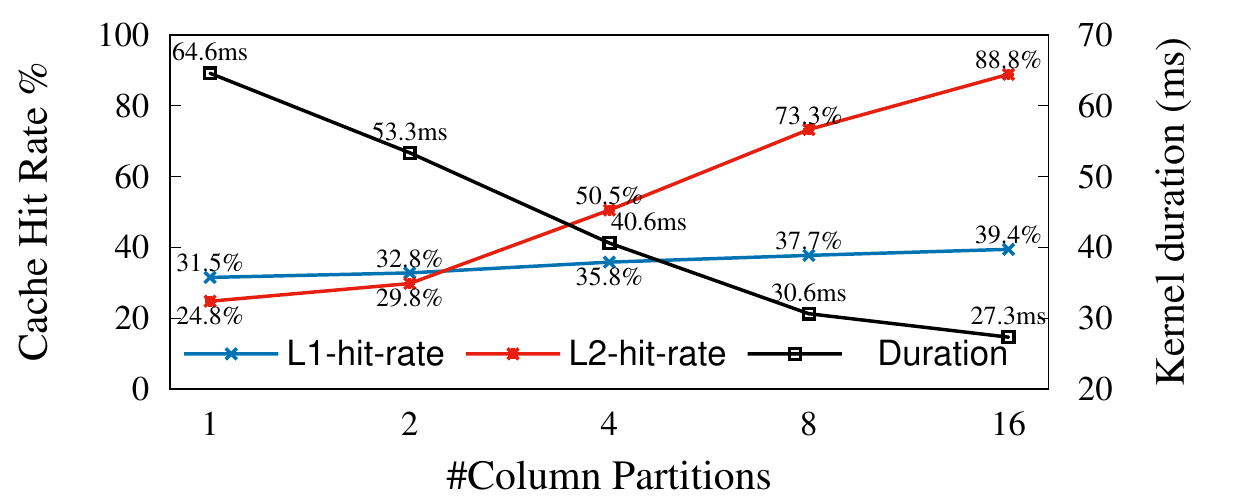}
    \caption{The kernel duration and L1/L2 hit-rate of \sys\ SpMM kernels under different column partitions.}
    \label{fig:kernel-metric}
\end{figure}

\begin{figure*}[ht]
    \centering
    \includegraphics[width=\textwidth]{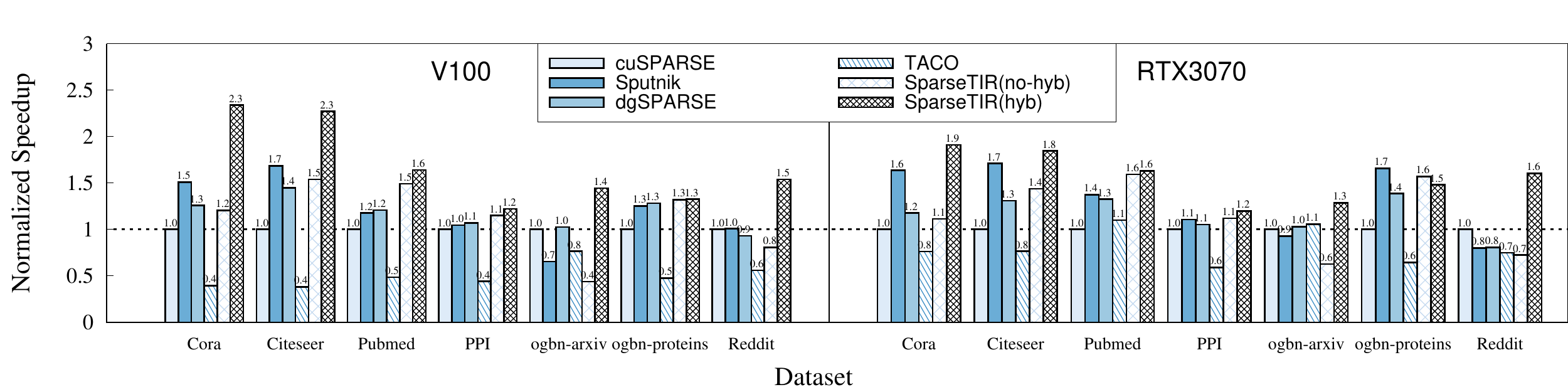}
    \caption{Normalized speedup against cuSPARSE for SpMM. \sys\  consistently outperforms vendor libraries and TACO. Comparing \sys(no-hyb) and \sys(hyb) demonstrates the importance of format composability.}
    \label{fig:spmm-eval}
\end{figure*}

We evaluate the SpMM written in \sys\ with and without the proposed $hyb$ format on real-world GNN datasets for both V100 and RTX3070. We measure the geometric mean speedup of different SpMM implementations against cuSPARSE for feature size $d\in\{32, 64, 128, 256, 512\}$. Figure \ref{fig:spmm-eval} shows our results. The \sys\ kernel on $hyb$ format obtains a 1.22-2.34x speedup on V100 and a 1.20-1.91x speedup on RTX 3070 compared to cuSPARSE. We also achieve consistently better performance than state-of-the-art open source sparse libraries dgSPARSE and Sputnik, and TACO with auto-scheduling enabled ~\cite{taco-sparse-iteration-Senanayake-2020}. Though TACO also explores compile-time load balancing, it does not support caching the partially aggregated result in registers, which is critical to kernel performance on GPUs, and the irregularity of the CSR format limits the application of loop unrolling. \sys\ perform these optimizations in stage II schedules.
\paragraph{Importance of composable formats.} We evaluate the \sys\ kernel without format decomposition (see \sys(no-hyb) in the figure). Results suggest that the \sys\ kernel without format decomposition and per-format scheduling performs generally worse: ogbn-arxiv is a citation network graph whose degrees obey power-law distribution, and our designed format can perform significantly better because of more efficient load balancing. {Notably, though padded zeros in our proposed composable format slightly increase FLOPs as shown in Table ~\ref{tab:datasets}, the runtime of \sys\ generated kernels on composable format is still faster because of better scheduling.} The degree distribution of the ogbn-proteins graph is centralized, and the benefit of using a  hybrid format is  compensated for the extra overhead introduced by padding. To evaluate the effect of column partitioning, we fix the feature size to 128 and measure several kernel metrics generated by \sys\ on a Reddit dataset under a different column partition setting. Figure \ref{fig:kernel-metric} shows the results;  L1 and L2's cache hit rates improve as we increase the number of column partitions. However, more partitions will increase the required memory transactions of the kernel because we will need to update the results matrix $c$ times if the number of partitions is $c$. As a result, the benefit of column partitioning saturates as we increase the number of partitions.

\subsubsection{SDDMM}
\label{sec:sddmm-eval}

SDDMM can be formulated as the following:

$$ B_{i,j} = \sum_{k=1}^{d} A_{i,j} X_{i, k} Y_{k, j}, $$

\noindent where $A$ and $B$ are two sparse matrices that share a  sparse structure, $X, Y$ are dense matrices, and $d$ is the feature size. In SDDMM, the computation per $(i, j)$ is independent, and the workload per position is the same, so we need not worry about load balancing issues if we parallelize the computation by each non-zero $(i, j)$. The \lstinline{sparse_fuse} schedule primitive in stage I introduced in Section \ref{sec:sparse-fuse} helps us iterate over non-zero $(i, j)$ directly instead of first iterating over $i$ and then iterating over non-zero $j$ for each $i$.

PRedS ~\cite{sddmm-yu-2021} is the state-of-the-art open-source SDDMM implementation, which optimizes SDDMM in two ways. First, it uses vectorized load/store intrinsics in CUDA,  such as \lstinline{float4}/\lstinline{float2}, which improves  memory  throughput. Second, it performs the reduction in two stages: (1) \textit{intra-group reduction}, which computes the reduction inside each group independently, and (2) \textit{inter-group reduction},  which summarizes the reduction result per group. We formulate the optimization in PRedS as composable transformations in \sys\ with \lstinline{vectorize} and \lstinline{rfactor} ~\cite{rfactor} schedule primitives at stage II, and we generalize the parameters, such as group size, vector length and number of workloads per CTA, as tunable parameters.

\begin{figure*}[ht]
    \centering
    \includegraphics[width=\textwidth]{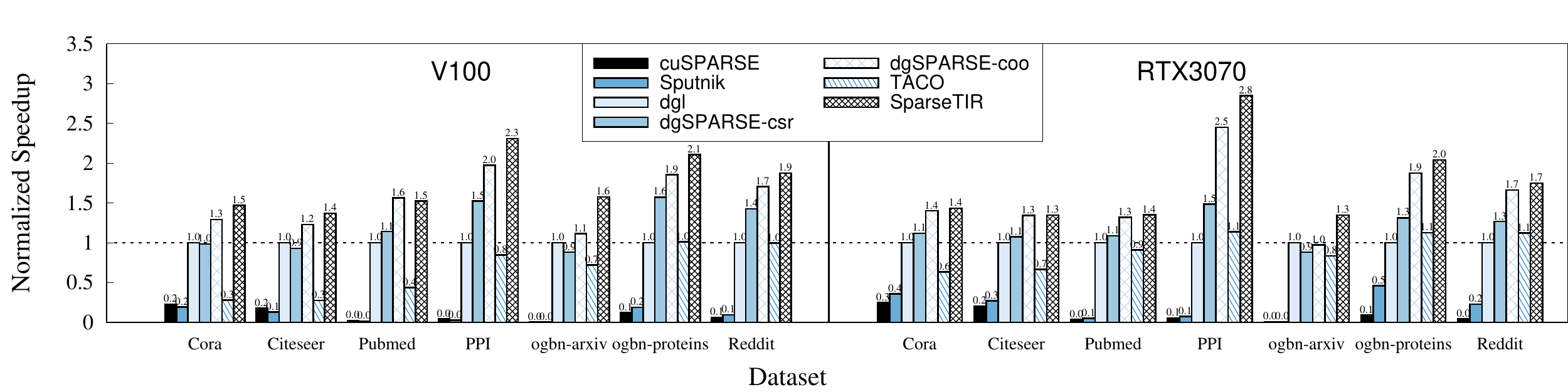}
    \caption{Normalized speedup against Featgraph for SDDMM. \sys\ beats the state-of-the-art vendor library dgSPARSE on average by parametrizing scheduling space.}
    \label{fig:sddmm-eval}
\end{figure*}

Figure \ref{fig:sddmm-eval} shows the geometric mean speedup of different SDDMM implementations vs our baseline for feature size $d\in\{32, 64,$ $128, 256, 512\}$. We do not use composable formats in SDDMM. The baseline we select is DGL's SDDMM implementation, which uses the optimization proposed in Featgraph ~\cite{featgraph-hu-2020}. cuSPARSE and Sputnik's SDDMM implementations are not optimized for highly sparse matrices such as graphs and thus achieve very low performance. We obtain generally better performance than dgSPARSE ~\cite{dgsparse},  which implements the PRedS ~\cite{sddmm-yu-2021} algorithm, because of the parameterized scheduling space. \sys\ significantly outperforms the DGL baseline and the 
TACO scheduled kernel because these implementations do not include two-stage reduction and vectorized load/store. 

\paragraph{Importance of composable transformations.} 
{The provenance graph data structure in TACO does not support multiple branches, thus we cannot perform schedules such as \lstinline{rfactor} at this level. The composable transformation design of \sys\ enables us to apply such schedules at lower stages.}

\subsubsection{End-to-end GraphSAGE training.}
\label{sec:graphsage-e2e-eval}

\begin{figure}[ht]
    \centering
    \includegraphics[width=0.3\textwidth, angle=270]{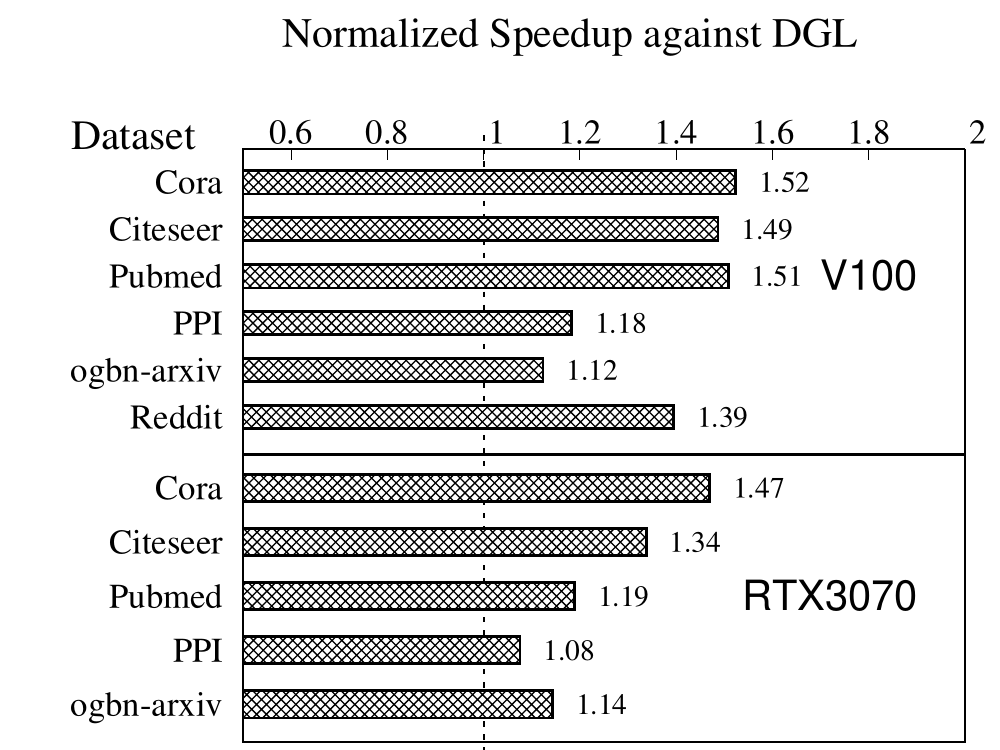}
    \caption{Normalized speedup of PyTorch+\sys\ against DGL on end-to-end GraphSAGE training.}
    \label{fig:graphsage-e2e-eval}
\end{figure}

We also integrate \sys-generated SpMM operators in the GraphSAGE~\cite{graphsage-hamilton-2017} model written in PyTorch and compare the end-to-end speedup to DGL. Figure  \ref{fig:graphsage-e2e-eval} shows that we obtain a 1.18-1.52x speedup on V100 and a 1.08-1.47x speedup on RTX 3070 \footnote{Reddit result is not reported on RTX 3070 because of Out-Of-Memory issue.}.

\subsection{Sparsity in Transformers}

{Sparsity in Transformers comes from (1) sparse attentions  ~\cite{sparsetransformer-rewon-2019, longformer-iz-2020, butterfly-beidi-2021}, and (2) sparsity in network weights after pruning ~\cite{movement-pruning, block-prunning}. We evaluate \sys\ generated kernel in both cases\footnote{In this section, we use half-precision data type for all operators to use Tensor Cores.}.}

\subsubsection{Sparse Attention.}
\label{sec:sparse-attention-eval}
\begin{figure}[ht]
    \centering
    \includegraphics[width=0.45\textwidth]{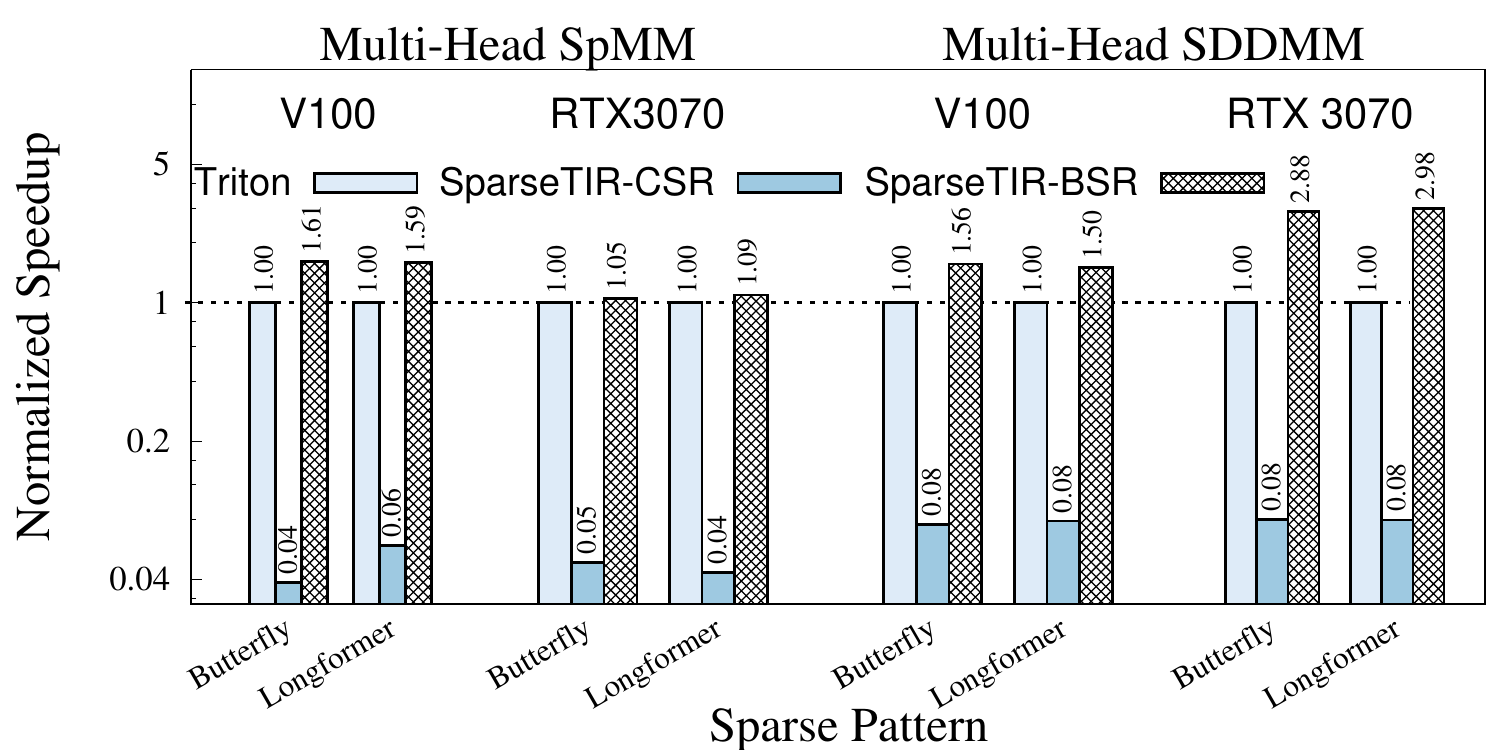}
    \caption{Normalized speedup against Triton on sparse transformer operators.}
    \label{fig:blocksparse}
\end{figure}

Sparse transformers reduce the complexity of Transformers by making the attention matrix sparse. The key operator in Sparse Transformers is still SpMM and SDDMM, but unlike GNNs whose sparse matrices are provided by graph structures, the sparse matrices used in sparse attentions are mostly manually designed and have a block-sparse pattern to better utilize tensor cores in modern GPUs. We select two examples: Longformer ~\cite{longformer-iz-2020} and Pixelated Butterfly Transformer ~\cite{butterfly-beidi-2021}, whose sparse structures are band matrix and butterfly matrix ~\cite{parker1995random}, respectively. We implement the batched-SpMM and batched-SDDMM operators for both CSR and BSR formats. For BSR operators, we use the \lstinline{tensorize} primitive during stage II IR schedules to use tensorized instructions in CUDA. Figure \ref{fig:blocksparse} shows different implementations' speedup against  Triton's ~\cite{triton-tillet-2019} block-sparse operator. We fix the matrix size to $4096\times4096$, batch(head) size to $12$, band size to $256$, and feature size per head to $64$. Results show that \sys-BSR obtains a 1.05-1.59x speedup on multi-head SpMM and a 1.50-2.98x speedup on multi-head SDDMM.

\subsubsection{Sparse Weight (Network Pruning)}

Network pruning ~\cite{prunning-han-2016} is another source of sparsity in Transformers. Pruning can significantly reduce the number of model parameters at the cost of negligible performance (accuracy) loss by making the weights sparse.
PruneBERT ~\cite{movement-pruning, block-prunning} applies pruning to Transformers, and we evaluate \sys's performance on PruneBERT in both structured pruning and unstructured pruning settings.

\paragraph{Structured Pruning.}
\label{sec:structured-pruning}

\begin{figure}[ht]
    \centering
    \includegraphics[width=0.45\textwidth]{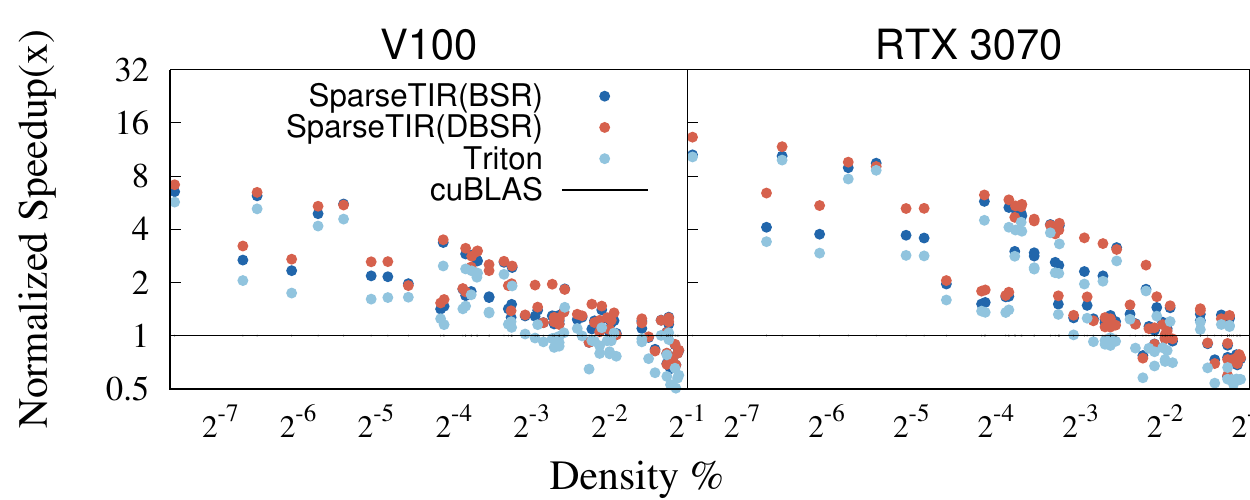}
    \caption{Normalized speedup against cuBLAS for operators extracted from block-pruned transformers. The $X$-axis refers to the weight density in the SpMM operator, and $Y$-axis refers to the normalized Speedup against cuBLAS implementation which uses a dense matrix for sparse weight.}
    \label{fig:structured-bert}
\end{figure}

{Structured Pruning prunes groups of weights together at the channel or block level to speed up execution. Block pruning ~\cite{block-prunning} is an example of structured pruning on Transformers where network weights are pruned to block-sparse format, the operator used in block-pruned Transformer is SpMM. We extract all SpMM operators in a block-pruned model\footnote{\url{https://huggingface.co/madlag/bert-base-uncased-squad1.1-block-sparse-0.07-v1}} with block size 32 and average weight sparsity 93\% for the benchmark. We fix the batch size to 1 and the sequence length to 512. Figure \ref{fig:structured-bert} shows the performance of \sys\ kernels, Triton's BSRMM, and cuBLAS on these operators. The block sparse weights in the block-pruned model have many all-zero rows, and we propose to use doubly-compressed BSR (DBSR, inspired by doubly compressed sparse row (DCSR) format ~\cite{dcsr}) format to skip zero rows. The results show that \sys\ kernel on DBSR format can consistently outperform \sys\ kernel on BSR format, and achieve better with Triton's BSRMM implementation.}

\paragraph{Unstructured Pruning.}
\label{sec:unstructured-pruning}

\begin{figure}[ht]
    \centering
    \includegraphics[width=0.4\textwidth]{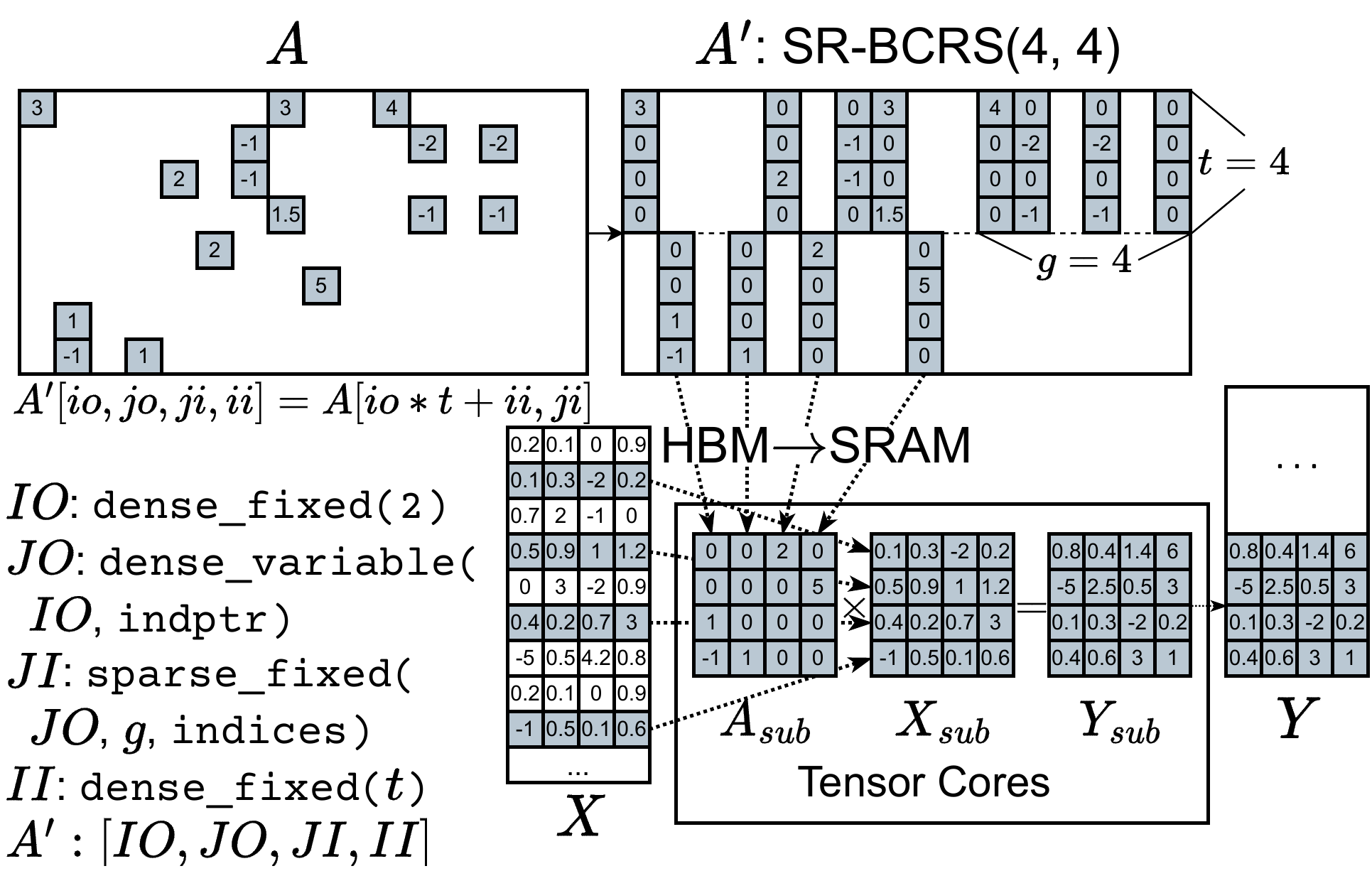}
    \caption{Conversion from unstructured sparse matrix to SR-BCRS$(t,g)$, and SpMM schedule on the it.}
    \label{fig:group-tile}
\end{figure}

{Unstructured pruning does not pose any constraints on the format of pruned weights, and the pruned weight matrices are typically stored in CSR format. Unstructured pruned model is known to be hard to optimize because of irregular computation, and directly converting them to BSR format would introduce too much fragmentation inside blocks. We use the SR-BCRS format proposed in Magicube ~\cite{magicube} to alleviate the issue. Figure \ref{fig:group-tile} explains how to represent SR-BCRS$(t, g)$ and corresponding SpMM schedules in \sys: the matrix is firstly divided into many $t\times 1$ tiles, and we omit tiles whose elements are all zero. The non-zero tiles inside the same rows are grouped by a factor of $g$, and we pad the tailing groups with zero tiles. Sparse matrices in SR-BCRS format can be composed by 4 axes in \sys. When performing SpMM on SR-BCRS, we can load a group of tiles in $A$ and corresponding rows in $X$ to local registers and use Tensor Cores in GPU (or Matrix-Multiply Units(MXU) in TPU ~\cite{tpu}, equivalently) to compute their multiplication results, these schedules can be described as \lstinline{cache-read/write} and \lstinline{tensorize} primitives at stage-II in \sys. Compared to BSR, the SR-BCRS format greatly reduces intra-block fragmentation: the non-zero ratio lower bound in SR-BCRS$(t, g)$ is $1/t$, while BSR with block size $b$ has a lower bound of $1/b^2$.}

\begin{figure}[ht]
    \centering
    \includegraphics[width=0.45\textwidth]{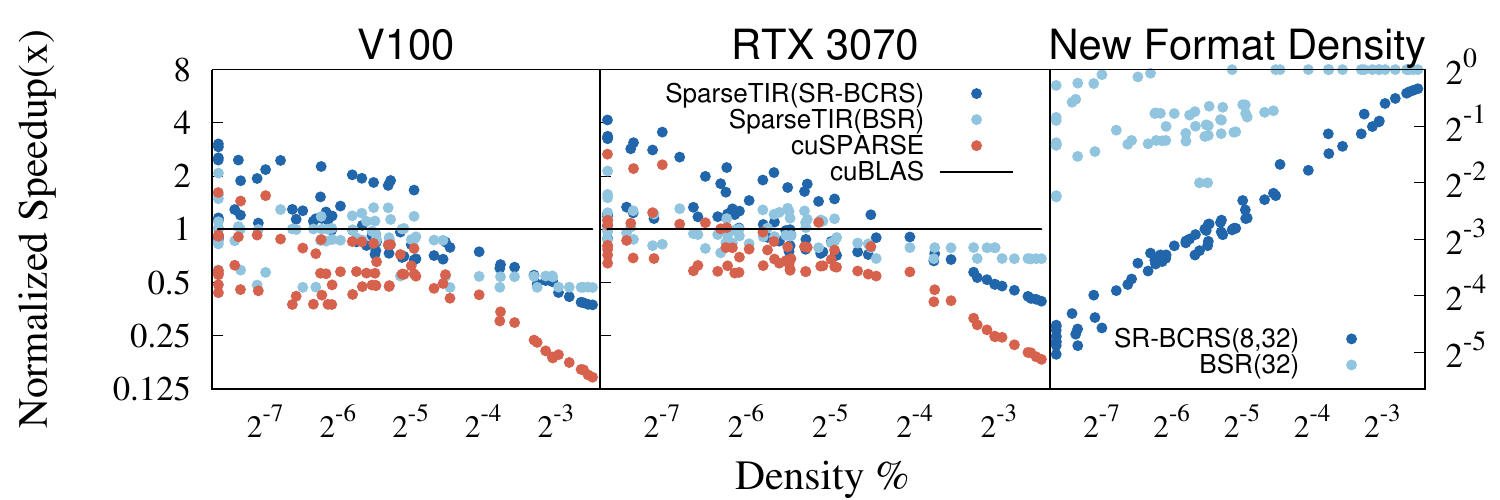}
    \caption{Normalized speedup aginst cuBLAS for operators extracted from unstructured pruned transformers, and the weight density in new format vs original weight.}
    \label{fig:unstructured-bert}
\end{figure}

{We extract all SpMM operators in a movement-pruned model\footnote{\url{https://huggingface.co/huggingface/prunebert-base-uncased-6-finepruned-w-distil-squad}} with average weight sparsity of 94\% for benchmark. Figure \ref{fig:unstructured-bert} shows the performance of \sys\ on SR-BCRS$(8, 32)$\footnote{To use m8n32k16 MMA instructions in GPU.}, BSR format with block size 32, and vendor libraries cuBLAS and cuSPARSE's CSRMM. We set the batch size to 1 and the sequence length to 512. We do not compare with Triton because it has no native implementation of SpMM on SR-BCRS. \sys\ on SR-BCRS beats \sys\ on BSR in most of the settings except for density $\geq 2^{-3}$, in which case both transformed sparse matrices have a density close to $1$. cuSPARSE's CSRMM can only beat cuBLAS' GeMM when weight density is extremely low (e.g., $\leq 2^{-6}$).}

\subsection{Relational Gather-Matmul-Scatter}

Relational Gather-Matmul-Scatter (RGMS for short) is an emerging sparse operator which can be expressed as follows:

$$Y_{i,l} = \sum_{r=1}^{R}\sum_{j=1}^{n}\sum_{k=1}^{d_{in}} A_{r,i,j} X_{j,k} W_{r,k,l}, $$
where $A$ is a 3D sparse matrix, whose leading dimension size is $R$, denoting number of relations. For each relation, the last two dimensions of $A$ form a unique 2D sparse matrix. $X$ is a 2D feature matrix and $W$ is a 3D weight matrix whose leading dimension size is also $R$. For each relation, the last two dimensions of $W$ form a unique 2D dense weight matrix. The scheduling for the RGMS operator is complicated because we need to consider (1) load balancing and (2) the utilization of Tensor Cores. Until now, no sparse library implements this kernel.

\subsubsection{Relational Graph Convolution Network.}
\label{sec:rgcn-eval}

{RGCN ~\cite{rgcn-schlichtkrull-2017} is a generalization of GCN model to heterogeneous graphs (graphs with multiple relations/edge types). The operator used in RGCN is RGMS, where $A_r$ refers to the adjacency matrix corresponding to sub-graph whose edge type is $r$, and $W_i$ refers to the weight matrix corresponding to edge type $r$. Table \ref{tab:rgcn-datasets} introduces the characteristics of heterogeneous graphs used in RGCN evaluation; in the table,
\#etypes refers to the number of edge types (also known as ``relations'') in the heterogeneous graph, \%padding refers to the ratio of padded zero elements after we transform the original sparse matrix with composable formats.} Existing GNN libraries implement RGMS operator in a two-stage approach:

\begin{align}
T_{r,j,l} & = \sum_{k=1}^{d_{in}} X_{j,k} W_{r,k,l}, \label{eq:rgms-1} \\
Y_{i,l} & = \sum_{r=1}^{R}\sum_{j=1}^{n} A_{r,i,j} T_{r,j,l}, \label{eq:rgms-2}
\end{align}

\begin{figure*}[ht]
    \centering
    \includegraphics[width=0.95\textwidth]{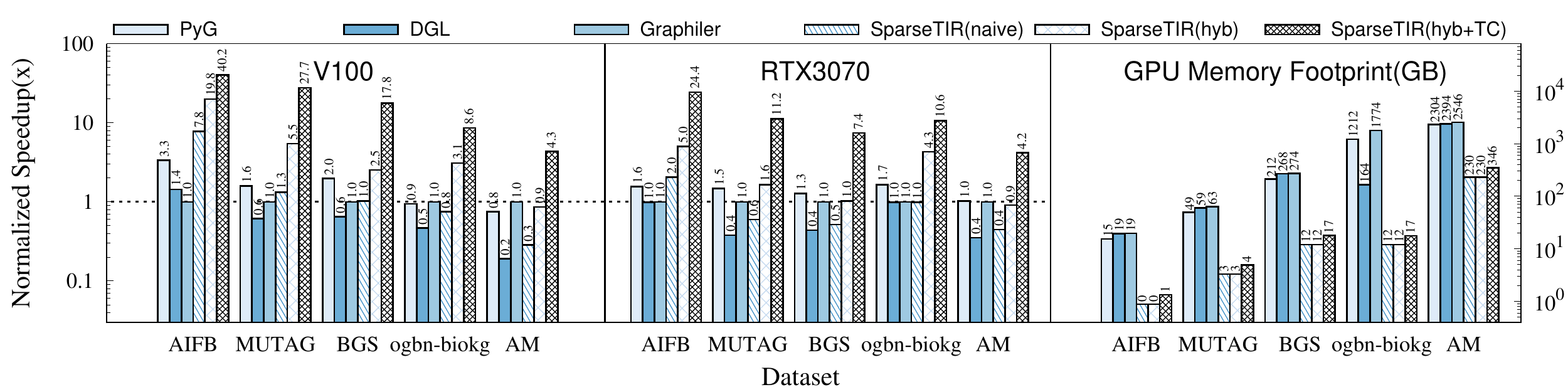}
    \caption{Normalized RGCN inference speedup against Graphiler and GPU Memory Footprint. \sys($hyb$+TC) uses schedule proposed in Figure \ref{fig:rgms-schedule}, \sys($hyb$) uses composable format but use CUDA Cores instead of Tensor Cores for on-chip Matrix Multiplication, \sys(naive) uses neither composable formats nor Tensor Cores.}
    \label{fig:rgcn-e2e-eval}
\end{figure*}

\begin{figure}[ht]
    \centering
    \includegraphics[width=0.45\textwidth]{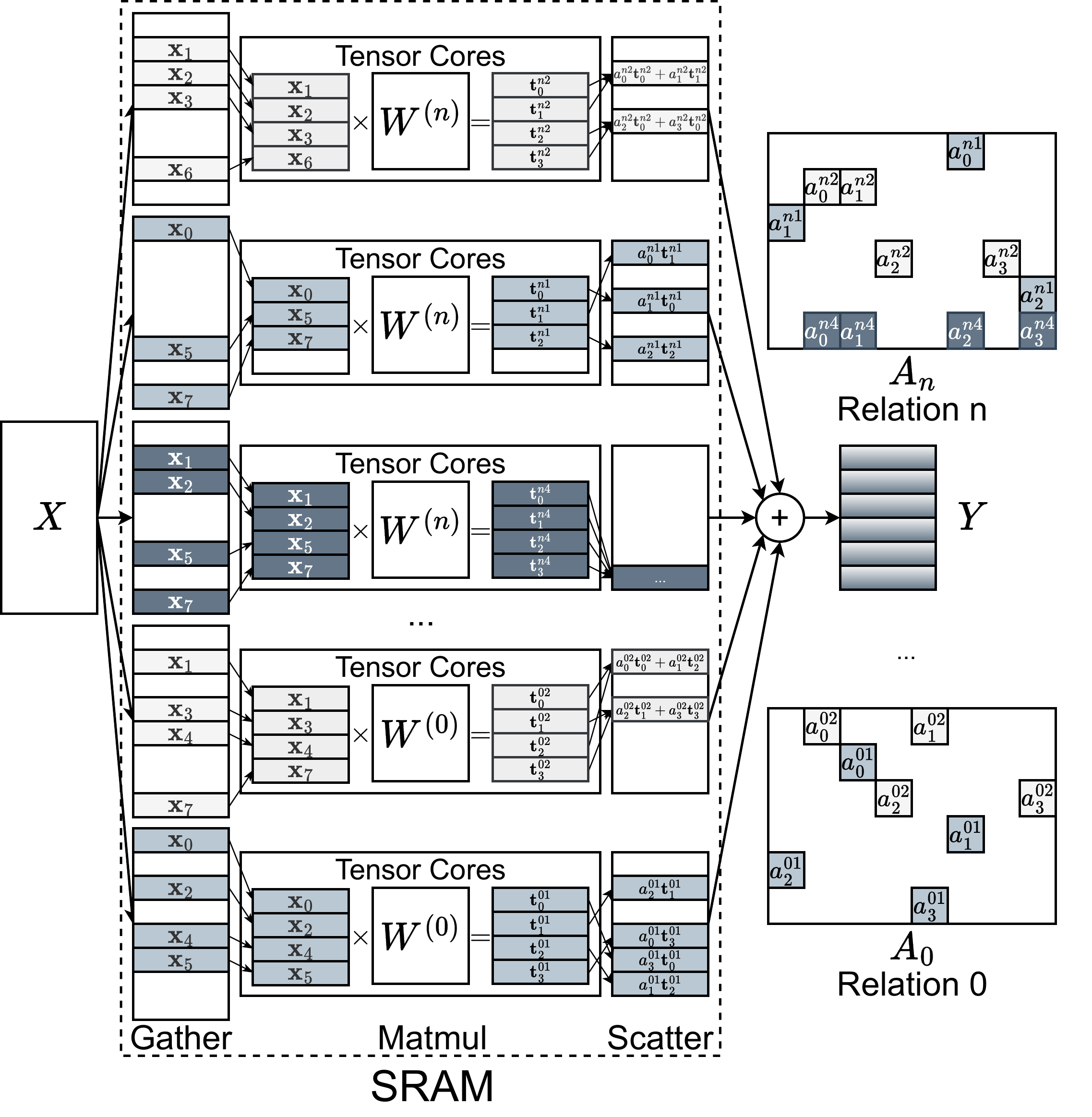}
    \caption{Schedule of RGMS operator in \sys. Composable formats $hyb$ are used for load balancing.}
    \label{fig:rgms-schedule}
\end{figure}

\begin{table}[t]
\centering
\begin{tabular}{lllll}
\toprule
Graph         & \#nodes   & \#edges & \#etypes & \%padding \\
\midrule
AIFB ~\cite{semantic-web}         & 7,262     & 48,810 & 45 & 17.9 \\
MUTAG ~\cite{semantic-web}        & 27,163    & 148,100 & 46 & 8.0 \\
BGS ~\cite{semantic-web}          & 94,806    & 672,884 & 96 & 4.3 \\
ogbl-biokg~\cite{hu2020ogb}    & 93,773    & 4,762,678 & 51 & 4.2 \\
AM ~\cite{semantic-web}           & 1,885,136 & 5,668,682 & 96 & 10.8 \\
\bottomrule
\end{tabular}
\caption{Statistics of Heterogeneous Graphs used in RGCN.}
\label{tab:rgcn-datasets}
\end{table}

where the first stage fuses gathering and matrix multiplication, and the second stage performs scattering. Such implementation materializes the intermediate result $T$ on HBM, which incurs a lot of GPU memory consumption. In \sys\ we fuses the two stage into a single operator: we generalize the $hyb$ format proposed in Figure \ref{fig:spmm-hyb-format} to 3-dimensional so that 2D sparse matrix corresponding to each relation is decomposed to $hyb(1, 5)$ formats. Figure \ref{fig:rgms-schedule} explain the scheduling of RGMS operator on 3D $hyb$ in \sys: for each ELL matrix $A^{rk}$ ($r$ refers to edge type and $k$ refers to bucket index), we pin its corresponding weight matrix $W^{r}$ in SRAM and gather related rows of $X$ from HBM to SRAM, then perform partial matrix multiplication with Tensor Cores and scatter results to $Y$. Note that the matrix multiplication and intra-group scatter are all performed inside SRAM. Such design reduces the overhead of data copy between SRAM and HBM for intermediate matrix $T$. We evaluate end-to-end RGCN inference (feature size: 32) and Figure \ref{fig:rgcn-e2e-eval} shows results: \sys($hyb$+TC) can significantly improve previous state-of-the-art GNN compiler Graphiler~\cite{graphiler-xie-2022} by 4.2-40.2x in different settings. By comparing \sys(naive), \sys($hyb$) and \sys($hyb$+TC) we show that both composable formats and composable transformations (which enables Tensorization) matter: even though $hyb$ increases FLOPs by padding zeros (as shown in Table ~\ref{tab:rgcn-datasets}), it still makes the kernel faster by 2-4.4x because of better load-balancing. \sys's generated fused kernel can also greatly reduce GPU memory footprint because we do not explicitly stores $T$ in HBM, with fragments of $T$ consumed immediately after produced in SRAM. \sys($hyb$+TC) consumes more GPU memory than \sys(naive) and \sys($hyb$) because of the half-precision/single-precision data type conversion.

\subsubsection{Sparse Convolution} 
\label{sec:sparse-conv-eval}

\begin{figure}[ht]
    \centering
    \includegraphics[width=0.45\textwidth]{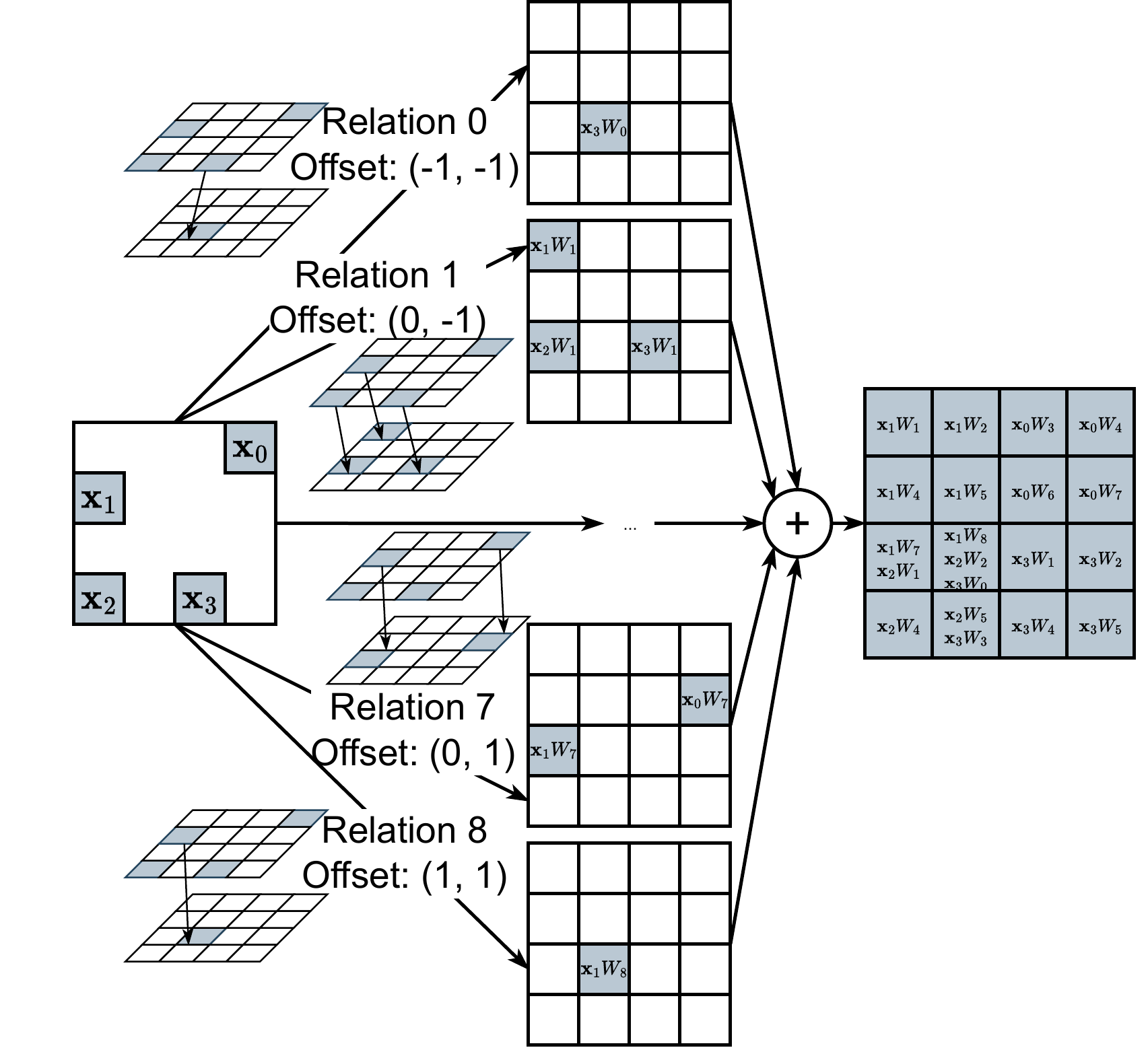}
    \caption{Equivalence of RGMS and Sparse Convolution, each relative offset inside the convolution kernel forms a relation in RGMS. The equivalence also holds in 3D setting.}
    \label{fig:rgcn-spconv}
\end{figure}

Sparse Convolution ~\cite{minkowskinet} is widely used in 3D cloud point data. We found that the Sparse Convolution operator is a special form of RGMS, and Figure \ref{fig:rgcn-spconv} illustrates the equivalence: each relative offset inside the convolution kernel can be viewed as a relation in RGMS. For each relation, the mapping between non-zero elements in feature map of previous layer to non-zero elements in feature of next layer forms a bipartite graph which can be viewed as a 2D sparse matrix whose number of non-zero elements per row is no greater than $1$.

\begin{figure}[ht]
    \centering
    \includegraphics[width=0.45\textwidth]{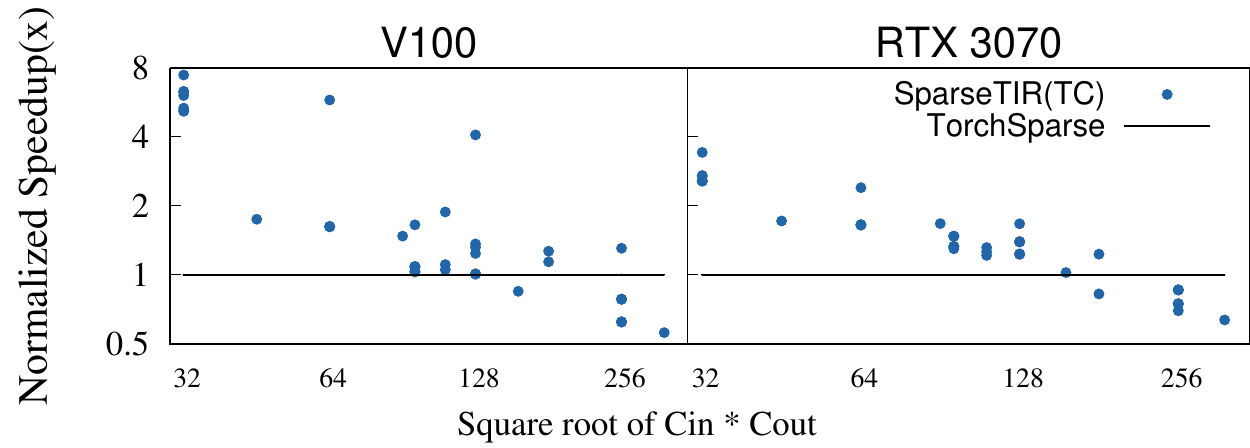}
    \caption{Normalized speedup against TorchSparse for Sparse Convolution. The $X$-axis refers to square root of input channel and output channel: $\sqrt{C_{in}C_{out}}$, and the $Y$-axis refers to speedup against TorchSparse.}
    \label{fig:sparse-conv-results}
\end{figure}

We extract all of the Sparse Convolution operators in MinkowskiNet ~\cite{minkowskinet} on SemanticKitti dataset ~\cite{semantickitti} for benchmark, and evaluate \sys's RGMS kernel\footnote{We don't need to use composable formats for Sparse Convolution because the sparse matrix for each relation is already an $ELL(1)$.}. Figure \ref{fig:sparse-conv-results} shows our normalized speedup against state-of-the-art TorchSparse ~\cite{torchsparse-tang-2022} library. Unlike the \sys's schedule in Figure \ref{fig:rgms-schedule}, TorchSparse does not fuse Gather-Matmul-Scatter on chip. Instead, it explicit materializes $T$ and uses coarse-grained cuBLAS operators rather than Tensor-Core level instructions for matrix multiplication\footnote{It's not necessary to use adaptive matrix multiplication grouping when using fine-grained Tensor-Core instructions.}. \sys's RGMS can outperform TorchSparse for most of the operators because of less HBM/SRAM data exchange as mentioned before. However, for large channel size ($> 128$), \sys's RGMS cannot beat TorchSparse because matrix multiplication overhead become dominant (The FLOPs of Matmul is quadratic to channel size while the FLOPs of Gather and Scatter is linear to channel size) and cuBLAS is better optimized than \sys's RGMS for large channel.

\section{Related Work}
\label{sec:related-work}

\paragraph{Tensor and deep learning compilers.} Halide ~\cite{halide-ragan-2013} and TVM ~\cite{tvm-chen-2018, autotvm-chen-2018} are tensor compilers that decouple kernel description and schedules for dense computation. XLA ~\cite{xla} and Relay~\cite{relay-jared-2018} proposed computational-graph-level abstractions for deep learning, where we can apply optimizations such as \textit{kernel fusion} and \textit{graph substitution} ~\cite{taso-jia-19}. However, these compilers have limited support for representing and optimizing sparse operators, impeding the wider deployment of sparse deep learning workloads such as GNNs. TensorIR ~\cite{tensorir} is TVM's new tensor-level programming abstraction for automatic tensorization. Triton ~\cite{triton-tillet-2019} is an intermediate language that offers tile-level operations and optimizations, FreeTensor ~\cite{freetensor} is a compiler for irregular tensor programs with loop-based programming model. These IRs could serve as stage-III IR for \sys.

\paragraph{Sparse compilers.}

MT1 ~\cite{aart-thesis, advanced-compiler-optimization-for-sparse-computations-aart-1995, auto-data-structure-selection-aart-1996, aart-ics-1993,nonzero-analysis-aart-1994}, SIPR ~\cite{pugh_sipr_1998}, Ironman ~\cite{mateev_next-generation_2000} and Ahmed et al. ~\cite{ahmed_framework_2000} introduces the idea of compiling kernels for a given sparse data structure and a kernel description. TACO ~\cite{taco-Kjolstad-2017, format-abstraction-chou-2018, kjolstad:2018:workspaces} proposes sparse format abstractions and a merge lattices-based code generation routine. Senanayake et al. ~\cite{taco-sparse-iteration-Senanayake-2020} propose a sparse-iteration space transformation framework for scheduling sparse operators. Chou et al. ~\cite{taco-conversion-2020} introduce an approach for generating  efficient kernels for sparse format conversion. Henry et al. ~\cite{sparse-compile-taco-henry-2021} generalize TACO to sparse array programming. These works have huge impact on the design of \sys. Sympiler ~\cite{cheshmi_sympiler_2017} builds a symbolic inspector to analyze sparse structure at compile-time and generates efficient code. Parsy ~\cite{cheshmi_parsy_2018} generalize the idea to support parallelization. SPF ~\cite{strout_sparse_2018} proposes a inspector-executor framework compatible with polyhedral transformations. Mohammadi et al. ~\cite{mohammadi_sparse_2019} proposes data dependence simplication algorithm for compiler generated inspectors.
These compilers have huge potential for utilizing sparse structures, and we're exploring the possibility of combining them with composable formats.
 Taichi ~\cite{taichi-hu-2019} decouple data structure and kernel description for physics simulation programming; its compiler optimization focuses on spatial sparse data, unsuitable for DL. Tiramisu ~\cite{tiramisu-sparse-2020} supports automatic selection of dense/sparse kernels at computational graph-level. However, it lacks tensor-level sparse code generation.
COMET ~\cite{comet-tian-2021} and MLIR Sparse Dialect ~\cite{sparse-mlir-aart-2022} are two MLIR dialects that explore composable IR design for sparse tensor algebra. Both treat sparse tensors with format annotation as first-class members in the IR; however, neither considers decomposable formats. CoRA ~\cite{cora-pratik-2021} proposes a compiler infrastructure for ragged tensors~\cite{ragged-tensor}: a special form of sparse tensors. The operation splitting in CoRA is also a special case of format decomposition in \sys. SparTA ~\cite{sparta} proposes sparse annotations for network pruning; its annotation is still dense and thus not applicable to highly sparse matrices used in GNNs. SparseLNR ~\cite{sparselnr} proposes \textit{branched iteration graph} to support factoring reductions and loop-fusion for sparse tensor algebra, these schedules can be formulated as stage-I schedules in \sys\ as we support branches in the IR.

\paragraph{GNN systems and compilers.}

PyG ~\cite{pyg-fey-2019} and DGL ~\cite{dgl-wang-2019} propose  programming interfaces for the programming message-passing ~\cite{mpnn-gilmer-2017} modules in GNN models. Both frameworks use vendor libraries and handwritten operators to accelerate specific message-passing patterns. Featgraph ~\cite{featgraph-hu-2020} optimizes generic GNN operators with TVM. However, it fails to support more operators because TVM lacks sparsity support. FusedMM ~\cite{fusedmm-rahman-2021} fuses SDDMM and SpMM operators, thus accelerating GNN training and saving GPU memory footprint. FusedMM can be described and optimized in \sys. Seastar ~\cite{seastar-yidi-2021} and Graphiler ~\cite{graphiler-xie-2022} compile user-defined message-passing functions to their intermediate representations (IR) and then optimize the IR and emit template-based, target-specific code: these templates still have limited expressiveness and cannot consider a wide range of the optimization space. \sys\ could serve as a backend for these GNN compilers. GNNAdvisor ~\cite{gnnadvisor-yuke-2021} proposes a CUDA template for GNN computations and uses graph characteristics to guide the performance tuning of GNN training. QGTC ~\cite{qgtc-wang-2021} and TC-GNN ~\cite{tc-gnn-2022} explore accelerating GNNs with TensorCores. Notably, the ``condensing'' technique proposed in TC-GNN is equivalent to SpMM on SR-BCRS format as shown in Section \ref{sec:unstructured-pruning}. The contribution of these papers is orthogonal to \sys.

\paragraph{Sparse kernel optimizations.}

Merge-SpMM ~\cite{merge-spmm-carl-2018}, ASpT ~\cite{aspt-hong-2019}, GE-SpMM ~\cite{ge-spmm-huang-2020}, Sputnik ~\cite{sputnik-gale-2020} and DA-SpMM ~\cite{spmm-heuristic-adaptability-2022} explore different schedule spaces for SpMM optimization on GPUs. We carefully examined the optimizations suggested in theses papers and propose a composable abstraction to unify them. OSKI ~\cite{oski-vuduc} is a library for auto-tuning sparse operators, with a focus on optimizing operators on cache-based, super-scalar architectures such as CPUs. However, OSKI do not support customizing sparse operators.

\paragraph{Sparse format optimizations.}

Pichon et al. ~\cite{reorder-blocking-2017} propose to reorder rows and columns in 2D sparse matrices to increase the block granularity of sparse matrices. Li et al. ~\cite{sparse-reorder-li-2019} study the problem of reordering sparse matrices to improve cache locality of operators on them. Mehrabi et al. ~\cite{ispass-sparse-reorder-2021} and Wang et al. ~\cite{gnnadvisor-yuke-2021} propose to reorder rows and columns of sparse matrices to accelerate SpMM on GPUs. These algorithms can act as pre-processing steps in \sys\ to discover efficient composable formats. 

\paragraph{Hardware-efficient algorithms.}

There have been a growing trend of sparsity in Deep Learning ~\cite{sparsity-torsten-2021}. To make better use of underlying hardware, researchers propose pruning algorithms with block-sparsity ~\cite{block-prunning} and bank-sparsity ~\cite{nm-sparsity-2021, bank-sparsity-cao-2019} to utilize acceleration units in GPUs, and ES-SpMM ~\cite{cache-sample-spmm-2021} for load balancing. \sys's composable abstractions can help researchers explore more complex sparse patterns with ideal performance on modern hardware.

\section{Future Work}
\label{sec:future-work}

\paragraph{Automatic scheduling}
\sys\ still requires users to specify schedule templates like they do for the first-generation of Halide and TVM.
The Halide auto-scheduler ~\cite{halide-autotuning-andrew-2019}, FlexTensor ~\cite{flextensor-size-2020}, Ansor ~\cite{ansor-lianmin-2020} and Meta-scheduler ~\cite{metaschedule-shao-2022} have been proposed to automatically generate schedule templates for dense tensor compilers. We expect these techniques would also prove helpful for sparse compilation. Searching for the optimal schedule is time consuming, Ahrens et al. ~\cite{taco-cost-model-ahrens-2022} propose an asymptotic cost model for sparse tensor algebra to narrow the schedule space of sparse kernels, which could also benefit our work. 

\paragraph{Automatic format decomposition}
In this paper we explore only manually designed format decomposition rules. We leave automatic format selection ~\cite{nonzero-analysis-aart-1994,auto-data-structure-selection-aart-1996} and decomposition for future work.

\paragraph{Dynamic Sparsity}
Some models ~\cite{moe-noam-2017, switch-transformer-2022,sparse-training-ampere} exhibit dynamic sparsity, where the position of non-zero elements changes overtime thus searching for best schedule for each matrix become impractical. DietCode ~\cite{zheng_dietcode_2022} proposes \textit{shape-generic} search space, micro-kernel based cost model and a lightweight dispatcher to dispatch kernel at runtime, the idea is also applicable to sparse tensor programs.

\paragraph{Integration with graph-level IR}
\sys\ models only tensor-level sparsity, we plan to extend the sparse attributes in \sys\ to graph-level IRs like XLA ~\cite{xla} and Relay ~\cite{relay-jared-2018}.

\section{Conclusion}
\label{sec:conclusion}

We introduce \sys, a composable abstraction for sparse operators in deep learning. Its key innovation is the use of composable formats and composable transformations, and together they form the parameter search space for performance tuning. Evaluations on generic sparse deep learning show that \sys\ achieves significant performance improvements over existing vendor libraries and frameworks.

\newpage
\begin{acks}
\label{sec:acknowledgements}

We thank all anonymous ASPLOS reviewers for their constructive comments. We thank Siyuan Feng, Bohan Hou and Wuwei Lin for discussions on tensorization and IR design, Sandy Kaplan for help on paper writing, Zhijian Liu for providing Sparse Convolution benchmarks, Joel Emer, Yuwei Hu, Jie Liu, Steven S. Lyubomirsky, Fredrik Kjosltad, Ye Tian, Zhiqiang Xie, and Zhongyuan Zhao for feedbacks on the paper.
This work was supported in part by the Center for Intelligent Storage and Processing in Memory (CRISP), a Semiconductor Research Corporation (SRC) program co-sponsored by DARPA. It was also supported by the Real Time Machine Learning (RTML) NSF and DARPA program, and the NSF award CCF-1518703, CNS-2211882.
The opinions and conclusions in this paper do not reflect the views of these funding agencies.

\end{acks}
\appendix

\section{Programming Interface for Composable Formats}

This section further explains the programming interface for composable formats and the format decomposition pass introduced in \S \ref{sec:format-decomposition},
\sys\ provide two APIs for composable formats:

\begin{description}
  \item[FormatRewriteRule] is a class for a sparse format rewriting rule description, its input include: the name of format rewrite rule, the sparse buffer to rewrite, a \sys\ description of new format, the mapping from original axes to new axes, and the index mapping $f$ and inverse index mapping $f^{-1}$ between original sparse buffer $A$ and the transformed sparse buffer $A'$: $A[\textbf{I}] = A'[f(\textbf{I})], A[f^{-1}(\textbf{I'})] = A'[\textbf{I'}]$, both $f$ and $f^{-1}$ need to be affine maps written in Python's lambda functions.
  \item[decompose\_format] is a function that accepts a list of format rewrite rules and an \sys\  program as input and performs the format decomposition pass on the given \sys\ program. 
\end{description}

\noindent Below is an example illustrating how to use the two APIs to compose \lstinline{ELL(2)} and \lstinline{BSR(2)} rewrite rules and perform format decomposition in Figure \ref{fig:format-decomposition}:

\begin{lstlisting}[language=tir]
@T.prim_func
def spmm(
    a: T.handle, b: T.handle, c: T.handle,
    indptr: T.handle, indices: T.handle,
    m: T.int32, n: T.int32, nnz: T.int32, feat_size: T.int32
) -> None:
    I = T.dense_fixed(m, idtype="int32")
    J = T.sparse_variable(
        I, (n, nnz), (indptr, indices), idtype="int32")
    J_ = T.dense_fixed(n, idtype="int32")
    K = T.dense_fixed(feat_size, idtype="int32")
    A = T.match_sparse_buffer(a, (I, J), "float32")
    B = T.match_sparse_buffer(b, (J_, K), "float32")
    C = T.match_sparse_buffer(c, (I, K), "float32")
    with T.sp_iter([I, J, K], "SRS", "csrmm") as [i, j, k]:
        with T.init():
            C[i, k] = 0.0
        C[i, k] = C[i, k] + A[i, j] * B[j, k]


def BSR(block_size: int):
    # block_size: the block size in BSR format.
    @T.prim_func
    def bsr_desc(
        a: T.handle,
        indptr: T.handle, indices: T.handle,
        m: T.int32, n: T.int32, nnz: T.int32
    ) -> None:
        IO = T.dense_fixed(m, idtype="int32")
        JO = T.sparse_variable(
            IO, (n, nnz), (indptr, indices), idtype="int32")
        II = T.dense_fixed(block_size, idtype="int32")
        JI = T.dense_fixed(block_size, idtype="int32")
        A = T.match_sparse_buffer(a, (IO, JO, II, JI), "float32")
        pass

    return FormatRewriteRule(
        "bsr_{}".format(str(block_size)),
        bsr_desc,
        ["A"], ["I", "J"], ["IO", "JO", "II", "JI"],
        {"I": ["IO", "II"], "J": ["JO", "JI"]},
        lambda i, j:
            return (i // block_size, j // block_size,
                    i % block_size, j % block_size)
        lambda io, jo, ii, ji:
            return io * block_size + ii, jo * block_size + ji
    )

def ELL(nnz_cols: int):
    # nnz_cols: number of non-zero columns per row in ELL format.
    @T.prim_func
    def ell(
        a: T.handle,
        indices: T.handle,
        m: T.int32, n: T.int32,
    ) -> None:
        I2 = T.dense_fixed(m, idtype="int32")
        J2 = T.sparse_fixed(
            I2, (n, nnz_cols), indices, idtype="int32")
        A = T.match_sparse_buffer(a, (I2, J2), "float32")
        pass

    return FormatRewriteRule(
        "ell_".format(str(nnz_cols)),
        ell_desc,
        ["A"], ["I", "J"], ["I2", "J2"],
        {"I": ["I2"], "J": ["J2"]},
        lambda i, j: return i, j
        lambda i2, j2: return i2, j2
    )

composable_format = [BSR(2), ELL(2)]
spmm_hybrid = decompose_format(spmm, composable_format)
\end{lstlisting}

\noindent where the prefix \lstinline{T} is used to prevent name conflicts with keywords in Python. Note that format conversion is a special case of format decomposition where we only put one \lstinline{FormatRewriteRule} in the list of composable formats.

\section{Artifact Appendix}

\subsection{Abstract}

This artifact includes scripts and dependencies for reproducing all experiments in the paper. We require a host with x86\_64 CPU and NVIDIA GPUs with Turing or later architectures to run the artifact. The \sys\ compiler is a submodule in the artifact, which is implemented in C++ and Python. The benchmarking is mainly written in Python. We modify the source code of some old dependencies to make sure they are compatible with the software version specified in the Dockerfile. We provide a docker image for users to run benchmarks inside the container, and scripts to generate tables and figures for comparison.

\subsection{Artifact check-list (meta-information)}

{\small
\begin{itemize}
  \item {\bf Data set: } OGB, SemanticKITTI, DGL built-in datasets.
  \item {\bf Run-time environment: } NVIDIA Container Toolkit.
  \item {\bf Hardware: } NVIDIA GPUs with Turing/Ampere/Hopper architecture.
  \item {\bf Execution: } All kernels being profiled are executed in GPUs, some data pre-processing are performed on CPUs.
  \item {\bf Metrics: } Execution time, GPU memory footprint.
  \item {\bf Output: } Execution time/GPU memory usage tables, and  figures.
  \item {\bf Experiments: } SpMM, SDDMM, GraphSAGE end-to-end training, Sparse Transformer operators, 3D Sparse Convolution, Relational Graph Convolutional Networks inference.
  \item {\bf How much disk space required (approximately)?: } 55GB.
  \item {\bf How much time is needed to prepare workflow (approximately)?: } 2 hour for building docker container.
  \item {\bf How much time is needed to complete experiments (approximately)?: } 10 hours.
  \item {\bf Publicly available?: } Yes.
  \item {\bf Code licenses (if publicly available)?: } The \sys-artifact is distributed under The MIT license and the \sys\ compiler is released under the Apache License, v2.0.
  \item {\bf Archived (provide DOI)?: } \url{https://doi.org/10.5281/zenodo.7643745}
\end{itemize}
}

\subsection{Description}

\subsubsection{How to access}

The artifact is available on Github: \url{https://github.com/uwsampl/sparsetir-artifact} and Zenodo: \url{https://doi.org/10.5281/zenodo.7643745}. Which includes the installation scripts for all dependencies and benchmark scripts to reproduce results. The SparseTIR compiler, which is available at \url{https://github.com/uwsampl/sparsetir}, has been incorporated as a submodule of the artifact.

\subsubsection{Hardware dependencies}
We conduct experiments on two machines, one with NVIDIA RTX 3070 GPU and another with NVIDIA Tesla V100 GPU, both of them are equipped with x86\_64 CPUs. Other NVIDIA GPUs with Turing, Ampere, or Hopper architecture should also work. A GPU with memory greater than or equal to 16GB is enough to reproduce all results, otherwise, users might encounter an Out-Of-Memory issue for relatively large datasets like \textit{Reddit} on end-to-end GraphSAGE training.

\subsubsection{Software dependencies}
We create a Docker image for this artifact, enabling users to run all experiments on a platform that meets the \href{https://docs.nvidia.com/datacenter/cloud-native/container-toolkit/install-guide.html}{installation requirements} of the NVIDIA Container Toolkit.

\subsubsection{Data sets}
For GNN-related experiments, we use Open Graph Benchmark ~\cite{hu2020ogb} and built-in datasets provided by DGL ~\cite{dgl-wang-2019}, for Sparse Convolution, we use SemanticKITTI dataset ~\cite{semantickitti}, for PrunedBERT, we use models publicly available in HuggingFace ~\cite{huggingface}.

\subsection{Installation}

To install the artifact, users can either clone the repository and build the artifact by themselves:
\begin{lstlisting}[language=bash]
  git clone https://github.com/uwsampl/sparsetir-artifact.git --recursive  
  cd sparsetir-artifact
  docker build -t sparsetir .
\end{lstlisting}

or pull the pre-built image we provided from the docker hub (only compatible with Ampere 
NVIDIA GPU architecture):
\begin{lstlisting}[language=bash]
  docker image pull expye/sparestir-ae:latest
  docker tag expye/sparsetir-ae:latest sparsetir
\end{lstlisting}

\subsection{Experiment workflow}

We provide a \lstinline{run.sh} script under each folder to run corresponding benchmarks:

\begin{description}
  \item[spmm] contains scripts to reproduce SpMM experiments in \S \ref{sec:spmm-eval}.
  \item[sddmm] contains scripts to reproduce SDDMM experiments in \S \ref{sec:sddmm-eval}.
  \item[e2e] contains scripts to reproduce GraphSAGE end-to-end training experiments in \S \ref{sec:graphsage-e2e-eval}.
  \item[sparse-attention] contains scripts to reproduce Sparse Transformer operator experiments in \S \ref{sec:sparse-attention-eval}.
  \item[pruned-bert] contains scripts to reproduce PrunedBERT experiments in \S \ref{sec:structured-pruning} and \S \ref{sec:unstructured-pruning}.
  \item[rgcn] contains scripts to reproduce RGCN inference end-to-end experiments in figure \S \ref{sec:rgcn-eval}.    
  \item[sparse-conv] contains scripts to reproduce Sparse Convolution operator experiments in \S \ref{sec:sparse-conv-eval}.
\end{description}

The scripts will produce logging files containing the profiling results including average execution time and GPU memory usage, and figures plotted in the same style as the paper. We use \lstinline{CUDAEvent} APIs to profile CUDA kernels. During profiling, we discard the samples for the first 10 runs as warm-up steps and repeat for 100 cycles.

\subsection{Evaluation and expected results}

The specific running time and speedup differ on different platforms but we expect the results users reproduced should roughly match the numbers reported in the paper. (see Figures \ref{fig:spmm-eval}, \ref{fig:sddmm-eval}, \ref{fig:graphsage-e2e-eval}, \ref{fig:blocksparse}, \ref{fig:structured-bert}, \ref{fig:unstructured-bert}, \ref{fig:rgcn-e2e-eval} and \ref{fig:sparse-conv-results}).

\subsection{Experiment customization}
Artifact users can customize the benchmark scripts to use other datasets, for GNN operator or end-to-end training/inference benchmarks, users can create their own datasets as \lstinline{DGLGraph} class (the graph data structure used in DGL). For the sparse convolution benchmark, users need to convert the customized point cloud dataset to \lstinline{SparseTensor} class introduced in TorchSparse. For the network pruning benchmark, user can convert their own pruned weights to scipy sparse matrix.

\subsection{Notes}
Many previous work do not flush L2 cache when profiling CUDA kernels, which results in incorrect measurement especially for ``small'' operators, because the data accessed in the previous run would reside in L2 cache thus reducing the memory latency in the next run if they are accessed before being replaced. In this artifact we provide an option for the user to determine whether to enable L2 or not: if the environment variable \lstinline{FLUSH_L2} is set to \lstinline{ON}, we enable L2 flush for all benchmarks, and if \lstinline{FLUSH_L2} is set to \lstinline{OFF} we will disable L2 flush. All experiment results reported in this paper are obtained with \lstinline{FLUSH_L2=ON}.

\bibliographystyle{ACM-Reference-Format}
\bibliography{./bib/papers.bib}

\end{document}